\documentclass{article}
\PassOptionsToPackage{numbers, compress}{natbib}
\usepackage[main,final]{neurips_2026}

\usepackage[utf8]{inputenc} 
\usepackage[T1]{fontenc}    
\usepackage{url}            
\usepackage{nicefrac}       
\usepackage{microtype}      

\usepackage{wrapfig}

\usepackage{amsmath}
\usepackage{amssymb}
\usepackage{mathtools}
\usepackage{amsthm}

\usepackage{hyperref}
\usepackage{graphicx}
\usepackage{multirow}
\usepackage{booktabs}
\usepackage{float}
\usepackage{makecell}
\usepackage{algorithm}
\usepackage{algorithmic}
\usepackage{enumitem}
\usepackage{xcolor}
\usepackage{diagbox} 
\usepackage{amsfonts}
\usepackage{bbding}
\usepackage{subcaption}
\usepackage{float}

\title{CoLD: Counterfactually-Guided Length Debiasing for Process Reward Models in Mathematical Reasoning}

\author{
    Congmin Zheng\thanks{Equal Contribution.} \textsuperscript{ \rm 1},
    Jiachen Zhu$^*$\textsuperscript{\rm 1},
    Jianghao Lin\textsuperscript{\rm 1},
    Xinyi Dai\textsuperscript{\rm 2},
    \\
    \textbf{Weiwen Liu}\textsuperscript{\rm 1},
    \textbf{Haoxuan Li}\textsuperscript{\rm 3},
    \textbf{Yong Yu}\textsuperscript{\rm 1},
    \textbf{Weinan Zhang}\thanks{Corresponding Author.} \textsuperscript{ \rm 1},
    \textbf{Mengyue Yang}\textsuperscript{\rm 4} \\
    \textsuperscript{\rm 1}Shanghai Jiao Tong University \textsuperscript{\rm 2}Huawei Noah’s Ark Lab\\\textsuperscript{\rm 3}Peking University \textsuperscript{\rm 4}University of Bristol \\
    desp.zcm@sjtu.edu.cn, gebro13@sjtu.edu.cn\\
    wnzhang@sjtu.edu.cn, mengyue.yang@bristol.ac.uk
}

\begin{document}

\maketitle

\begin{abstract}
Process Reward Models (PRMs) play a central role in evaluating and guiding multi-step reasoning in large language models (LLMs), especially for mathematical problem solving. However, we identify a pervasive length bias in existing PRMs: a tendency to assign higher scores to more verbose reasoning steps, regardless of their semantic content or logical validity. This bias undermines the reliability of reward predictions and leads to overly verbose outputs during inference. To address this issue, we propose \textbf{CoLD} (\textbf{Co}unterfactually-Guided \textbf{L}ength \textbf{D}ebiasing), a unified framework that mitigates length bias based on counterfactual reasoning and causal graph analysis through three components: (1) an explicit length-penalty module, (2) a trainable bias estimator to capture spurious length-related signals, and (3) a joint training strategy that disentangles semantic correctness from superficial length features. Extensive experiments on MATH500 and GSM-Plus show that CoLD improves accuracy in step selection, and encourages more concise, logically valid reasoning. Furthermore, it consistently improves downstream RL performance and generalizes across domains by mitigating length bias, demonstrating CoLD’s strong generalization capability.
\end{abstract}

\section{Introduction}
\label{sec:Introduction}

\begin{wrapfigure}{R}{0.5\textwidth}
  \vspace{-15pt} 
  \begin{subfigure}[t]{0.48\linewidth}
    \centering
    \includegraphics[width=\linewidth]{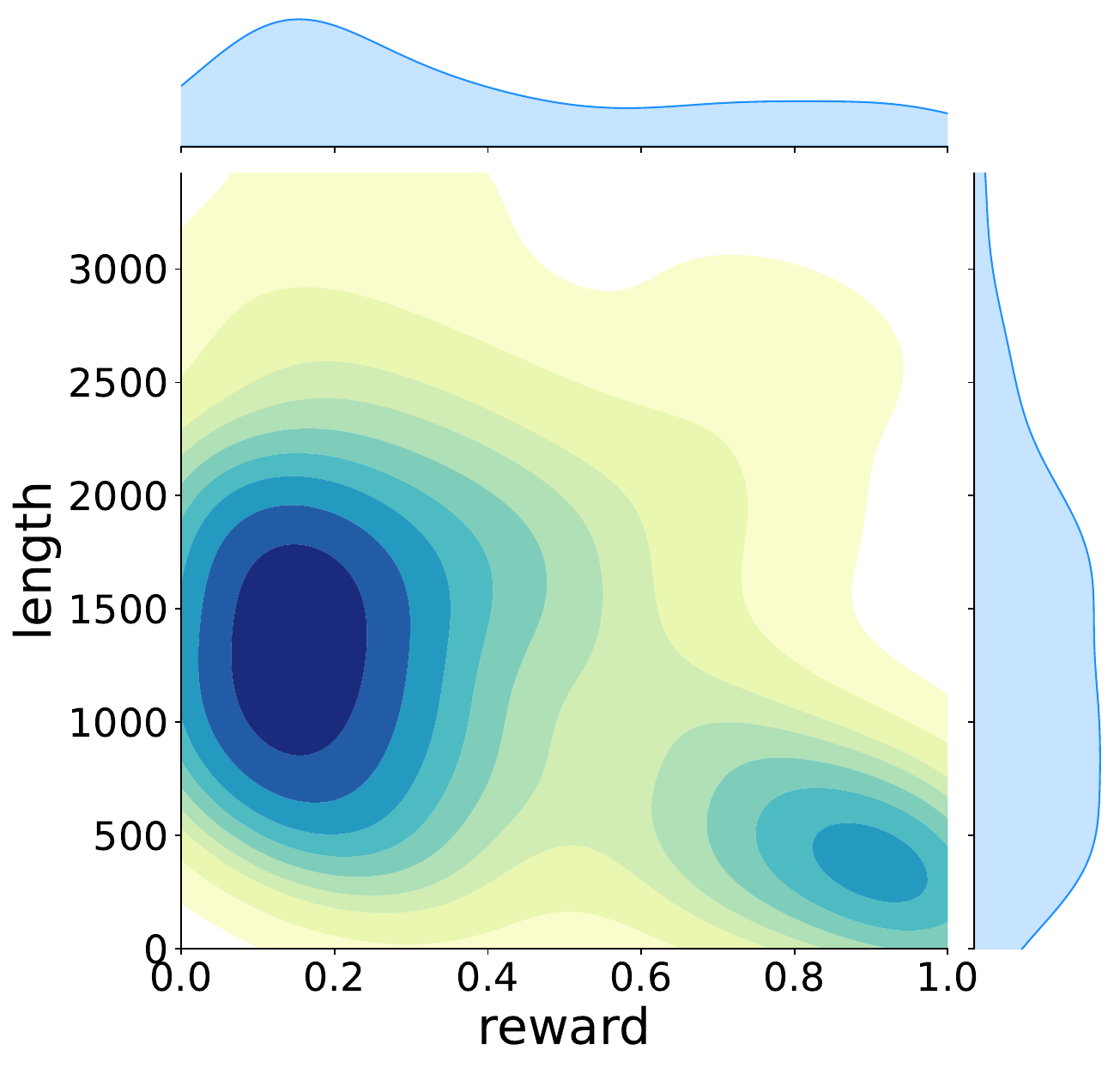}
    \caption{Original Version}
  \end{subfigure}
  \hfill
  \begin{subfigure}[t]{0.48\linewidth}
    \centering
    \includegraphics[width=\linewidth]{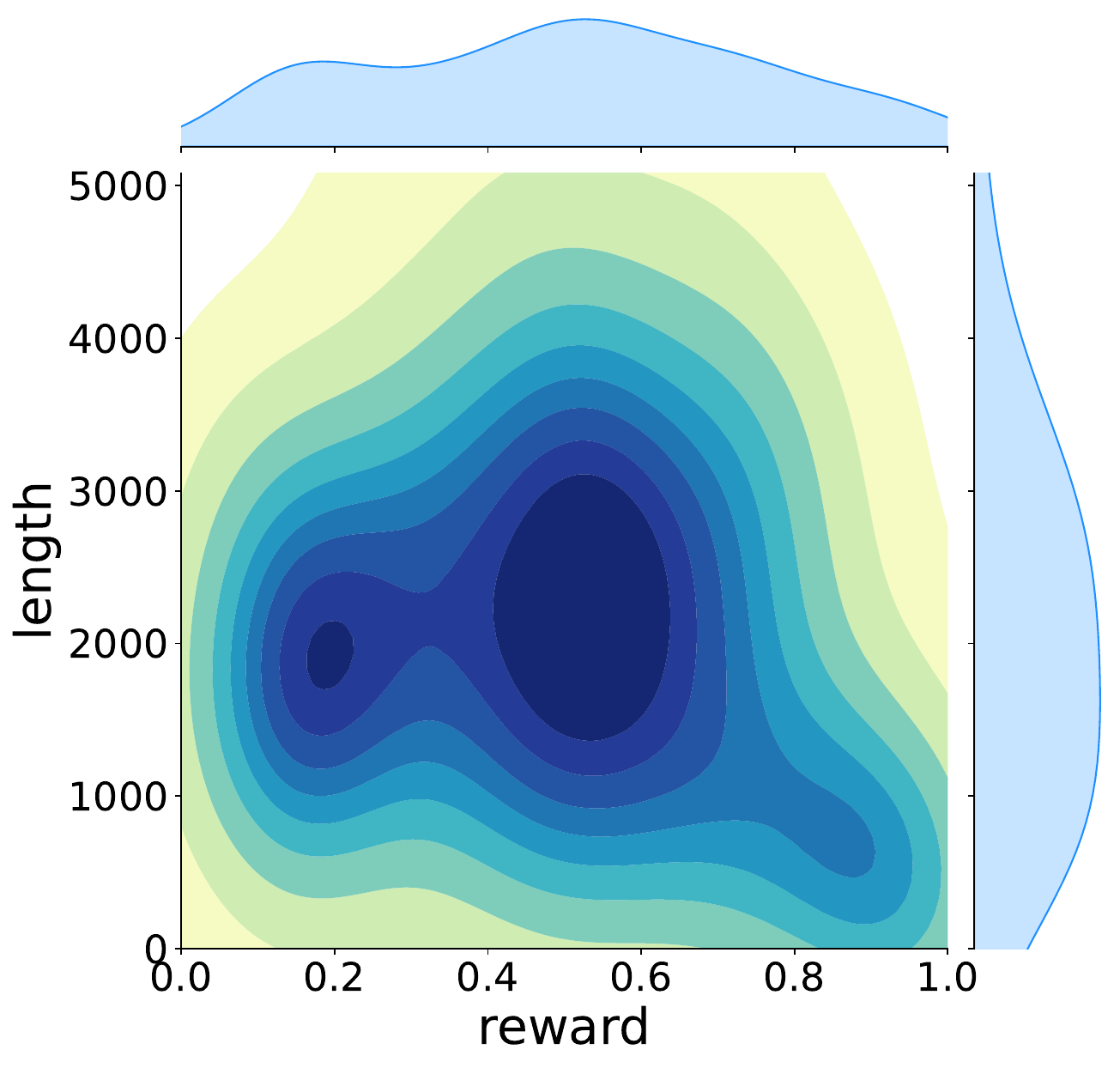}
    \caption{Extend Version}
  \end{subfigure}
  \caption{
  The joint distribution of reward score (x-axis) and step length (y-axis). (a) Distribution for the original reasoning steps. (b) Distribution for the same steps after being paraphrased by DeepSeek-V3 for increased textual length and verbosity, while maintaining semantic and logical equivalence.
  }
\label{fig:reward-length}
  \vspace{-10pt}
\end{wrapfigure}

Large language models (LLMs) have shown strong mathematical reasoning ability~\citep{openai2023gpt,dubey2024llama,zhu2024deepseek,shao2024deepseekmath,liu2024deepseek,yang2025qwen3}, yet their solutions often contain hidden reasoning errors despite yielding correct final answers~\citep{lightman2023let}. Process Reward Models (PRMs)~\citep{lightman2023let,wang2024math,zheng2025survey} were introduced to evaluate the logical soundness of intermediate steps and are now widely used in both error diagnosis and inference-time scaling strategies~\citep{wu2024inference,snell2024scaling,zhao2025genprm}.  Despite the impressive performance of recent PRMs~\citep{zhu2025retrieval,wang2024openropensourceframework,skyworkopeno12024,zou2025reasonflux}, we observe that their reward predictions can be unduly influenced by superficial features of reasoning steps, particularly their step-level textual length. 
We term this phenomenon \textbf{length bias}: \textit{the tendency of PRMs to assign higher scores to textually longer or more verbose steps, even if their semantic content and logical correctness are identical to more concise counterparts.}
Such bias undermines the reliability of PRMs, as it confuses surface-level verbosity with the quality of substantive reasoning.

\begin{figure}[t]
  \centering
  \vspace{-5pt}

  \includegraphics[width=\textwidth]{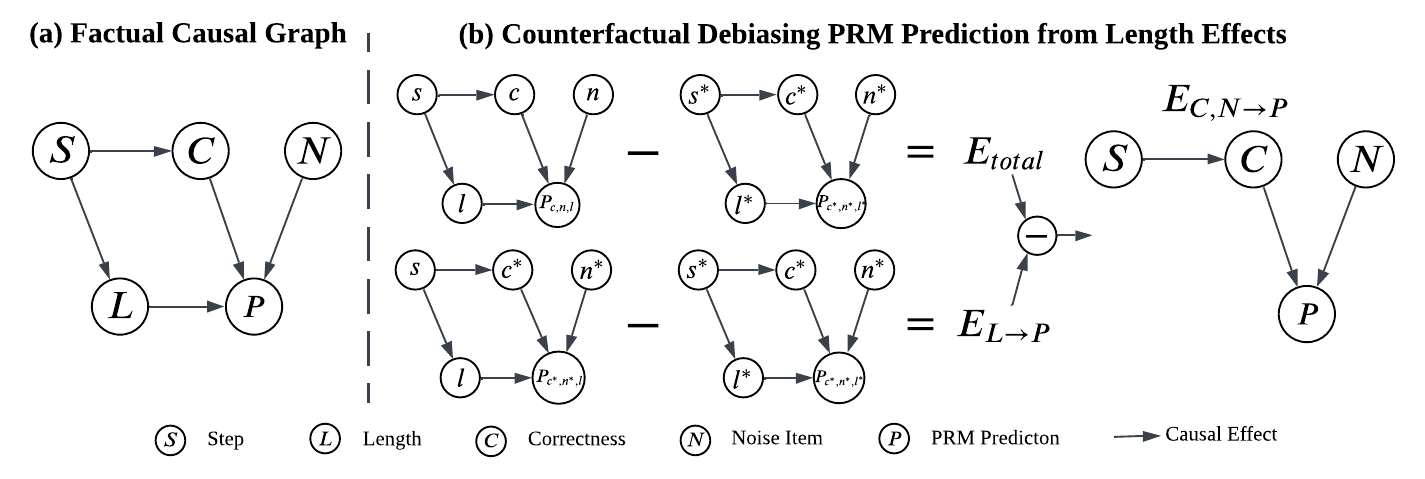}
\vspace{-10pt}
  \caption{Causal Modeling to Eliminate Length Bias in Prediction with Causal Graph of Factors Affecting PRM Reward. Uppercase letters (e.g., $L$) represent random variables, lowercase letters (e.g., $l$) denote specific values, and lowercase letters with an asterisk (e.g., $l^*$) indicate fixed values. $E$ denotes the causal effect.}
  \label{fig:causal graph}
\vspace{-15pt}
\end{figure}

To investigate this, we designed a controlled experiment decoupling step length from reasoning quality. Based on existing datasets~\citep{li2024process}, we constructed a semi-synthetic dataset where steps are extended via duplication or verbose LLM rewriting, strictly preserving their original semantics and correctness. We then input both the original and extended versions into the same PRM, recording the reward scores and token lengths. As shown in Figure~\ref{fig:reward-length}, the textually longer steps consistently receive higher rewards despite being logically equivalent to the originals. This suggests that PRMs exploit step length as a spurious shortcut, inflating scores in ways that undermine their reliability in assessing true reasoning quality.

To further analyze the factors influencing PRM predictions, we construct a causal graph~\citep{pearl2009causality} that captures the qualitative relationships among key variables, as illustrated in Figure~\ref{fig:causal graph}(a). The graph includes five nodes: the input step ($S$), length ($L$), logical correctness ($C$), latent noise factors ($N$), and PRM prediction ($P$). Ideally, predictions should follow the path $S \rightarrow C \rightarrow P$, ensuring that scores reflect reasoning quality. However, our analysis reveals a spurious path $S \rightarrow L \rightarrow P$, indicating that verbosity directly manipulates the reward—independent of logical value. While latent noise factors such as fluency or problem type ($N$) may also impact $P$, their influence appears secondary. Additional analysis is provided in the Preliminary section~\ref{sec:Causal-driven Analysis}.

Addressing this issue, we adopt a counterfactual perspective. We define length bias as the undesired change in a PRM’s output when step length varies while semantics and correctness are held fixed. This perspective motivates the goal of eliminating the score component attributable solely to length, as shown in Figure~\ref{fig:causal graph}(b). Guided by this insight, we propose \textbf{Co}unterfactually-Guided \textbf{L}ength \textbf{D}ebiasing (CoLD), a unified framework comprising three strategies: (1) Length Penalty: A simple adjustment subtracting a length-proportional term from the score; (2) Bias Estimator: an auxiliary model trained to estimate and remove length-induced bias to restore length invariance; (3) Joint Training: a unified training scheme that jointly optimizes the PRM and the Bias Estimator. By introducing input perturbations and regularization, the model is encouraged to disentangle semantic content from superficial length features.

Together, CoLD forms a principled debiasing strategy. By explicitly modeling, estimating, and removing spurious length effects, CoLD reduces the correlation between reward and verbosity while maintaining semantic fidelity. Empirically, it improves selection accuracy and promotes concise, logically sound responses.

Our main contributions are as follows:
\begin{itemize}[leftmargin=10pt]
    \vspace{-5pt}
    \item We are the first to identify and empirically validate length bias in PRMs, showing the tendency of PRMs to favor verbose steps over equally correct concise ones. Through causal graph analysis, we further reveal that step length is a confounding factor that distorts reward estimation and undermines the reliability of PRMs.
    \vspace{-5pt}
    \item We introduce the CoLD framework, grounded in counterfactual reasoning, to mitigate length bias in PRMs. CoLD significantly improves the fairness and accuracy of reward assessments.
    \vspace{-5pt}
    \item Extensive experiments show that CoLD improves step-selection accuracy and reasoning conciseness, while consistently enhancing downstream RL performance and cross-domain generalization through length-bias mitigation.
    \vspace{-5pt}
\end{itemize}

\section{Preliminary}
\label{sec:preliminary}
\subsection{Formulation}

We consider the problem of assigning reward scores to intermediate reasoning steps in mathematical problem solving. Let $q$ denote a math question and $\mathbf{s} = \{s^1, s^2, \ldots, s^n\}$ be a sequence of solution steps. For each prefix $x^j = (q, s^{\le j})$ comprising the first $j$ steps, a PRM assigns a scalar reward:
\begin{equation}
    r(x^j) \in (0,1),
\end{equation}
which is intended to reflect the semantic correctness of the partial solution up to step $j$.

To implement $r(x^j)$ using LLMs, we follow a scoring protocol that computes the reward based on the model’s classification between correctness and incorrectness. Specifically, given $x^j$, we extract the logits of two special answer tokens (e.g., '+' for correct and '-' for incorrect), and apply a softmax to compute a score:
\begin{equation}
    r(x^j) = \frac{\exp(l_+)}{\exp(l_+) + \exp(l_-)},
\end{equation}
where $l_+$ and $l_-$ denote the logits assigned to the positive and negative options, respectively.

\subsection{Causal-driven Analysis}
\label{sec:Causal-driven Analysis}
Causal graphs~\citep{pearl2009causality} provide a principled framework for modeling complex dependencies. We leverage this perspective to analyze PRM reward prediction and inform our framework design.

Figure~\ref{fig:causal graph}(a) illustrates the causal graph underlying PRM reward prediction, each node represents a causal factor, and each directed edge $A \rightarrow B$ indicates that $A$ exerts a direct causal influence on $B$.



\begin{itemize}[leftmargin=10pt]
\vspace{-5pt}
    \item The causal graph consists of five nodes: $S,L,C,N$ and $P$. $S$ denotes the given problem-solving step, $L$ represents the length of the given step, $C$ represents the logical correctness of the problem-solving step. $N$ represents latent noise factors such as linguistic fluency. $P$ represents  PRM's reward prediction for this step. 
    \vspace{-3pt}
    \item For a given step, it determines both length and correctness, resulting in causal edges $S \rightarrow L$ and $S \rightarrow C$. While $L$ and $C$ have no direct causal link, a confounding relation may exist. For example, extremely short steps often fail to convey valid reasoning, reducing the chance of correctness. 
    \vspace{-3pt}
    \item Ideally, $P$ should depend only on correctness $C$, i.e., $S \rightarrow C \rightarrow P$, making PRM a reliable evaluator that assigns high rewards to logically valid steps. In practice, however, the alternative path $S \rightarrow L \rightarrow P$ also plays a substantial role. Step length $L$ directly inflates the reward ($P$) independent of reasoning quality. This indicates the PRM exploits verbosity as a shortcut.
\vspace{-3pt}
    \item Finally, while unobserved noise $N$ may also directly influence $P$, as also shown in previous work~\cite{shen2023loose}, its effect is comparatively minor.
    \vspace{-5pt}
\end{itemize}

To isolate reasoning quality ($S \rightarrow C \rightarrow P$) from spurious length influence ($S \rightarrow L \rightarrow P$), we formulate a counterfactual objective: ensuring rewards remain invariant to length while holding correctness fixed, as illustrated in Figure~\ref{fig:causal graph}(b). A more detailed theoretical analysis is presented in Appendix~\ref{sec:Causal Counterfactual Analysis}. Based on the above analysis, we introduce a framework to eliminate length bias.

\section{Methodology}
\label{sec:method}

\begin{figure}[t]
  \centering
\vspace{-5pt}
  \includegraphics[width=1.0\textwidth]{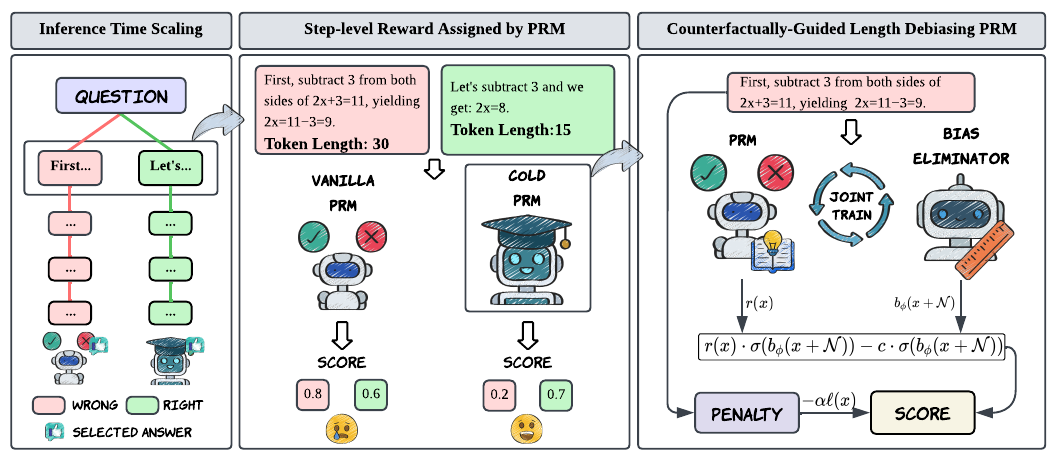}
\vspace{-5pt}
  \caption{Overview of the Counterfactually-Guided Length Debiasing (CoLD) framework for Process Reward Models, in comparison to the vanilla PRMs.}
  \label{fig: Framework}
\vspace{-15pt}

\end{figure}

\subsection{Overview of CoLD PRM}
\label{sec:overview}
As illustrated in Figure~\ref{fig: Framework}, CoLD PRM guides the selection of high-quality reasoning paths by systematically eliminating length bias from step quality evaluations. Rather than a single strategy, the framework integrates three complementary methods: (1) an explicit penalty term to discourage verbosity; (2) a learnable bias estimator isolating the length-based reward component; and (3) a joint optimization scheme synchronizing the PRM and the bias estimator.

Under this framework, the final debiased reward $r^*(x)$ is computed as:
\begin{equation}
\label{eq:inference}
\begin{split}
    r^*(x) =&\ r_{\theta}(x) \cdot \sigma(b_\phi(x + \mathcal{N})) - c \cdot \sigma(b_\phi(x + \mathcal{N})) - \alpha \ell(x),
\end{split}
\end{equation}
where $r_{\theta}(x)$ denotes the score assigned by the PRM, $b_\phi(x + \mathcal{N})$ denotes the output of the bias estimator given the noise-injected input $x + \mathcal{N}$ ($\mathcal{N}$ denotes the Gaussian noise following~\cite{shen2023loose}), $\sigma(\cdot)$ is the sigmoid function, $c$ is a hyperparameter controlling the strength of bias correction, and $\alpha \ell(x)$ represents the length penalty term. 

Crucially, this formulation is a direct algebraic realization of our causal counterfactual analysis (detailed in Appendix~\ref{sec:Causal Counterfactual Analysis}). The core objective of counterfactual debiasing is to isolate the pure semantic effect by subtracting the bias induced by length from the total prediction. In our framework, the factual prediction $P(C = c, L = \ell)$ is estimated using the composed reward $r_\theta(x)\sigma(b_\phi(x + \mathcal{N}))$. Conversely, the counterfactual baseline $P(C = c^*, L = \ell)$, which represents the spurious score component driven purely by length when true semantic correctness is counterfactually removed, is jointly approximated by both the neural estimator $\sigma(b_\phi(x + \mathcal{N}))$ and the structural prior $\alpha \ell(x)$. By executing this explicit subtraction during inference, CoLD successfully neutralizes the spurious $S \rightarrow L \rightarrow P$ pathway and recovers the debiased semantic reward.

Further analytical insights into this formulation are provided in the following discussion.

\subsection{Length Penalty}
While neural networks are highly capable of modeling bias, they can exhibit unstable predictions or saturation when encountering out-of-distribution, excessively long reasoning steps. To provide a robust safeguard, we first introduce the Length Penalty item $\alpha l(x)$. This term serves as a deterministic structural prior that explicitly penalizes verbosity by subtracting a length-proportional term from the original PRM score.

Here, $\alpha > 0$ is a hyperparameter controlling the penalty strength, and $\ell(x)$ denotes the token-level length of the reasoning step $x$. This formulation introduces an explicit dependency on length into the reward computation, encouraging concise responses and discouraging unnecessary verbosity. 

\subsection{Bias Estimator}
Length bias in PRMs is driven not merely by absolute token counts, but also by implicit, non-linear stylistic features (e.g., redundant phrasing or verbose transitions) that spuriously correlate with high rewards. To capture these complex verbosity patterns, we introduce a learnable module $b_\phi(x)$, designated as the Bias Estimator. The primary goal of the Bias Estimator is to flexibly model and eliminate the spurious reward components attributable to these verbose styles, thereby isolating superficial bias from the true semantic reward.

To prevent the estimator from absorbing actual reasoning semantics, we apply feature-space noise perturbation. Specifically, the input is defined as $\hat{x} = x + \mathcal{N}$, where $\mathcal{N}$ denotes injected Gaussian noise~\citep{shen2023loose} (Implementation details and analysis are in Appendix~\ref{app:noise}).  This design discourages $b_\phi$ from relying on semantic content and guides it to concentrate on surface-level factors such as length.

Together, the Bias Estimator and the Length Penalty form a complementary defense: the former targets non-linear features, while the latter enforces a reliable global boundary on absolute length.

\subsection{Joint Training of PRM with Bias Estimator}
Furthermore, we adopt a joint training strategy to simultaneously optimize the PRM $r_{\theta}(x)$ and the Bias Estimator $b_\phi(x)$. To disentangle correctness from stylistic features (e.g., length), we incorporate complementary correlation constraints.

We enforce complementary correlation constraints to ensure effective disentanglement between semantic correctness and length. Specifically, we measure the Pearson correlation between each module's output and the step length $\ell(x)$:
\begin{equation}
\rho_r = \frac{\mathrm{Cov}(r_{\theta}(x), \ell)}{\sigma_r \sigma_\ell} = \frac{\mathbb{E}\left[(r_{\theta}(x) - \mathbb{E}[r_{\theta}(x)])(\ell - \mathbb{E}[\ell])\right]}{\sigma_r \sigma_\ell},
\end{equation}
\begin{equation}
\rho_b = \frac{\mathrm{Cov}(b_\phi(x), \ell)}{\sigma_b \sigma_\ell} = \frac{\mathbb{E}\left[(b_\phi(x) - \mathbb{E}[b_\phi(x)])(\ell - \mathbb{E}[\ell])\right]}{\sigma_b \sigma_\ell},
\end{equation}
where $\mathrm{Cov}(\cdot, \cdot)$ denotes empirical covariance, and $\sigma_r$, $\sigma_b$, and $\sigma_\ell$ are the standard deviations. A small $\rho_r$ indicates that the PRM has reduced its reliance on length, while a large $\rho_b$ confirms that the Bias Estimator has effectively captured the length-dependent component. The corresponding correlation losses are defined as:
\begin{align}
\mathcal{L}_{\text{PRM}}({\theta}) &= \lambda_r \cdot \rho_r^2, \\
\mathcal{L}_{\text{Bias}}(\phi) &=- \lambda_b \cdot \rho_b^2,
\end{align}
where $\lambda_r$ and $\lambda_b$ control the strength of regularization. This asymmetric formulation promotes semantic fidelity in the PRM, while allowing the Bias Estimator to isolate and model spurious stylistic factors such as length.

However, relying solely on correlation-based constraints may lead to degenerate solutions that overlook semantic correctness. To ensure the debiased reward can still distinguish correct from incorrect steps, we add a cross-entropy loss as an additional supervision signal.

The composed reward is defined as $\hat{r}(x) = r_\theta(x) \cdot \sigma(b_\phi(\hat{x}))$, and is trained using correctness labels $y \in \{0, 1\}$ through a cross-entropy objective:
\begin{equation}
    \mathcal{L}_{\text{CE}}(\theta,\phi) = -\mathbb{E}_{(x, y) \sim \mathcal{D}} \Big[ y \log \sigma(\hat{r}(x)) +  (1 - y) \log (1 - \sigma(\hat{r}(x))) \Big],
\end{equation}
where $\mathcal{D}$ denotes the training data distribution. This loss encourages the composed reward to align with the ground-truth correctness signal.

The final training objective combines all components:
\begin{equation}
\mathcal{L}_{\text{Final}}(\theta,\phi)=\mathcal{L}_{\text{CE}}(\theta,\phi)+\mathcal{L}_{\text{PRM}}(\theta)+\mathcal{L}_{\text{Bias}}(\phi).
\end{equation}

\paragraph{Why training and inference use different formulations.}
This design addresses a fundamental challenge in the field of causal inference: the absence of ground-truth labels for counterfactual data. In real-world training datasets, the observed correctness labels $y$ are influenced by the confounding variable of inherent "verbosity," meaning they strictly belong to the observational factual data. Practically, we can't observe a counterfactual version of the same reasoning step where its correctness is removed while its surface length is held unchanged. Consequently, the counterfactual baseline required by the theoretical formula is completely unattainable during the training phase.

Therefore, our strategy is to separate learning from debiasing. During the training stage, we deliberately retain the biased structure within the composed reward $\hat{r}(x)$ so that it completely fits the factual distribution $P(C=c, L=\ell,N=n)$ that inherently contains confounding factors. Built upon this, the joint optimization framework ensures that the internal sub-networks accurately partition and model this biased signal. In the inference stage, we then apply Eq.~\ref{eq:inference} to perform an explicit counterfactual subtraction in a purely algebraic manner, thereby deriving a length-debiased reward that isolates semantic quality from superficial verbosity.

Further explanation of the algorithm is provided in Appendix~\ref{appendix:alg-desc}.

\paragraph{Alternative Strategy.}
In addition to joint training, an alternative strategy is to train the Bias Estimator independently while keeping the PRM fixed. This approach offers greater flexibility and modularity, especially when modifying existing PRMs without re-training the whole model.

\section{Experiments}
\label{sec:experiments}
\begin{table*}[t]
\centering    
\vspace{-10pt}
\caption{Performance comparison between CoLD and baseline RL debiasing methods. Results are reported as best-of-16 accuracy. The best and second-best results are highlighted in \textbf{bold} and \underline{underlined}, respectively. * denotes significant improvement over baseline (paired t-test, $p < 0.05$)}
\vspace{-5pt}
\label{tab:baseline performance}
\resizebox{\textwidth}{!}{
\renewcommand\arraystretch{1.2}
\begin{tabular}{clcccccc}
\hline
\multirow{2}{*}{\textbf{Policy Model}}&\multirow{2}{*}{\quad\textbf{Debias Method}} & \multicolumn{2}{c}{\textbf{MATH500}} & \multicolumn{2}{c}{\textbf{GSM-Plus}} & \multicolumn{2}{c}{\textbf{Avg}}  \\ 
\cmidrule(r){3-4} \cmidrule(r){5-6} \cmidrule(r){7-8}
 \multicolumn{2}{c}{} & \textbf{ArithACC(\%)} & \textbf{Length} & \textbf{ArithACC(\%)} & \textbf{Length}& \textbf{ArithACC(\%)} & \textbf{Length}\\ 
\hline 

\multicolumn{1}{c|}{\multirow{8}{*}{\makecell{Llama-3-\\70B-Instruct}}}&Vanilla-Base-PRM & 44.8 & 555.7 & 71.0 & 324.6 & 57.9 & 440.1 \\
\multicolumn{1}{c|}{\multirow{4}{*}{}}& \quad+ Length Penalty& 44.8&495.1& 71.5 & 300.1&58.2 & 397.6 \\
\multicolumn{1}{c|}{\multirow{4}{*}{}}& \quad+  Loose Lips Sink Ships& 45.8&435.7& \underline{72.5} & 	302.5 &59.2 & 368.9 \\
\multicolumn{1}{c|}{\multirow{4}{*}{}}& \quad+  Adaptive Length Bias Mitigation& 	43.0&	594.8& 	72.2 & 	364.3&57.6 & 479.6 \\
 \multicolumn{1}{c|}{\multirow{4}{*}{}}&\quad+ Uniform Average& 40.4&608.5& 70.4 & 324.6&55.4 & 466.6 \\
 \multicolumn{1}{c|}{\multirow{4}{*}{}}&\quad+ Locally Weighted Regression& 39.0&590.7&69.6 & 299.9&47.2 & 445.3 \\
\multicolumn{1}{c|}{\multirow{4}{*}{}}& \quad+ CoLD(w/o Joint) (Ours)& \underline{48.0}$^*$ & \underline{370.4}$^*$ & 72.1 & \underline{258.6}$^*$ & \underline{60.1}$^*$ & \underline{314.5}$^*$ \\
 \multicolumn{1}{c|}{\multirow{4}{*}{}}&\quad+ CoLD (Ours)& \textbf{49.2}$^*$ & \textbf{313.2}$^*$ & \textbf{73.8}$^*$ & \textbf{202.5}$^*$ &\textbf{61.5}$^*$ & \textbf{257.9}$^*$ \\

\hline

\multicolumn{1}{c|}{\multirow{8}{*}{\makecell{MetaMath-\\Mistral-7B}}}&Vanilla-Base-PRM & 37.0 & 555.2 & 59.5 & 335.5 & 48.2 & 445.4 \\
\multicolumn{1}{c|}{\multirow{4}{*}{}}& \quad+ Length Penalty& 37.6&448.8&59.1&313.9&48.4 & 381.4 \\
\multicolumn{1}{c|}{\multirow{4}{*}{}}& \quad+  Loose Lips Sink Ships&	34.5&		441.2&	58.0 & 		340.0&46.3& 390.6 \\
\multicolumn{1}{c|}{\multirow{4}{*}{}}& \quad+ Adaptive Length Bias Mitigation& 	31.0&	610.0& 	56.5 & 		385.0&43.8 &497.5 \\
\multicolumn{1}{c|}{\multirow{4}{*}{}}&\quad+ Uniform Average& 33.2&609.9&56.3 &363.8&44.8 & 486.9 \\
\multicolumn{1}{c|}{\multirow{4}{*}{}}&\quad+ Locally Weighted Regression& 29.6&507.3&52.7 & 354.4&41.2 & 430.9 \\
\multicolumn{1}{c|}{\multirow{4}{*}{}}& \quad+ CoLD(w/o Joint) (Ours)& \textbf{38.6}$^*$ & \textbf{353.4}$^*$& \underline{59.8} & \underline{262.7}$^*$ & \underline{49.2}$^*$ & \underline{308.1}$^*$ \\
\multicolumn{1}{c|}{\multirow{4}{*}{}}&\quad+ CoLD (Ours)& \underline{37.2} & \underline{376.3}$^*$ & \textbf{61.4}$^*$ & \textbf{238.6}$^*$ &\textbf{49.3}$^*$ & \textbf{307.5}$^*$ \\

\hline

\multicolumn{1}{c|}{\multirow{8}{*}{\makecell{Muggle-\\Math-13B}}}&Vanilla-Base-PRM & 30.4 & 411.1 & 59.1 & 287.9 & 44.8 & 349.5\\
\multicolumn{1}{c|}{\multirow{4}{*}{}}& \quad+ Length Penalty& 30.0&376.1&59.3 & 280.5&44.7 & 328.3 \\
\multicolumn{1}{c|}{\multirow{4}{*}{}}& \quad+ Loose Lips Sink Ships& 28.6&	375.7&	59.1 & 	263.9&43.9 & 319.8 \\
\multicolumn{1}{c|}{\multirow{4}{*}{}}& \quad+ Adaptive Length Bias Mitigation& 25.4&		472.0& 56.3 & 	364.5&40.9 & 418.3 \\
\multicolumn{1}{c|}{\multirow{4}{*}{}}&\quad+ Uniform Average& 27.4&440.8& 55.3 & 328.7&41.4 &384.8 \\
\multicolumn{1}{c|}{\multirow{4}{*}{}}&\quad+ Locally Weighted Regression& 23.5&470.5&50.7 & 312.3&37.1 &391.4 \\
\multicolumn{1}{c|}{\multirow{4}{*}{}}& \quad+ CoLD(w/o Joint) (Ours)& \underline{31.0}$^*$ & \underline{329.0}$^*$ & \underline{59.9}$^*$ & \textbf{238.3}$^*$ & \underline{45.5}$^*$ & \underline{283.7}$^*$ \\
\multicolumn{1}{c|}{\multirow{4}{*}{}}&\quad+ CoLD (Ours)& \textbf{31.4}$^*$& \textbf{309.2}$^*$ & \textbf{60.3}$^*$ & \underline{243.3}$^*$ & \textbf{45.9}$^*$ & \textbf{276.3}$^*$\\

\hline
\vspace{-15pt}
\end{tabular}
}
\end{table*}

In this section, we present the experimental settings
and results. Our code implementation is available in the repository\footnote{\url{https://anonymous.4open.science/r/CoLD-PRM-B640/}}.

\subsection{Experiment Setup}
\label{exp:setup}
\paragraph{Datasets and Metrics}
Our models are trained on the human-annotated PRM800K~\citep{lightman2023let} and the auto-labeled Math-Shepherd~\citep{wang2024math}, adopting established evaluation protocols to assess model.

For evaluation, we use the BON@n metric~\citep{lightman2023let,wang2024math} to assess the PRM’s ability to select the most correct reasoning trajectory from $n$ candidates. For each question, the PRM assigns scores to individual steps, and the overall score of a trajectory is determined by the minimum score among its constituent steps, following prior work~\citep{wang2024math}. Details about datasets can be found in Appendix~\ref{app:Implementation Details}.

\paragraph{Baselines and Implementation Details}
We consider a range of baselines and base models. Additional details, including further introductions to the baselines and base models, as well as hyperparameters and training sizes, are provided in Appendix~\ref{app:Implementation Details}.
\begin{table*}[t]
\centering
\vspace{-10pt}
\caption{Generalization of CoLD using the Llama-3-70B-Instruct. Performance is evaluated using Best-of-16 accuracy. $\uparrow/\downarrow$ indicate increase/decrease in accuracy/length. * denotes significant improvement over baseline (paired t-test, $p < 0.05$). Additional results are available in Table~\ref{tab:basemodel_metamath},~\ref{tab:basemodel_muggle}.}
\vspace{-5pt}
\label{tab:basemodel_llama3}
\resizebox{\textwidth}{!}{
\renewcommand\arraystretch{1.2}
\begin{tabular}{lcccccc}
\hline
\multirow{2}{*}{\textbf{PRM Model}} & \multicolumn{2}{c}{\textbf{MATH500}} & \multicolumn{2}{c}{\textbf{GSM-Plus}} & \multicolumn{2}{c}{\textbf{Avg}}  \\ 
\cmidrule(r){2-3} \cmidrule(r){4-5} \cmidrule(r){6-7}
 & \textbf{ArithACC(\%)} & \textbf{Length} & \textbf{ArithACC(\%)} & \textbf{Length}& \textbf{ArithACC(\%)} & \textbf{Length}\\ 
\hline 
Vanilla-Math-Shepherd-PRM & 46.4 & 458.9 & 72.3 & 268.7 & 59.4 & 363.8 \\
CoLD(w/o Joint)-Math-Shepherd-PRM & 47.2$\uparrow^*$ & 332.4$\downarrow^*$ & 72.5$\uparrow$ & 258.0$\downarrow^*$ & 59.9$\uparrow$ & 295.2$\downarrow^*$ \\
Vanilla-Qwen2.5-Math-PRM & 45.0 & 595.2 & 73.9 & 284.3 & 59.5 & 439.8 \\
CoLD(w/o Joint)-Qwen2.5-Math-PRM & 49.0$\uparrow^*$ & 330.6$\downarrow^*$ & 74.4$\uparrow^*$ & 217.6$\downarrow^*$ & 61.7$\uparrow^*$ & 274.1$\downarrow$ \\
Vanilla-EurusPRM-Stage1 & 49.0 &546.6 & 73.5 & 308.2 & 61.3 & 427.4 \\
CoLD(w/o Joint)-EurusPRM-Stage1 & 50.6$\uparrow^*$ & 282.1$\downarrow^*$ & 73.8$\uparrow$ & 237.7 $\downarrow^*$& 62.2$\uparrow^*$ & 259.9$\downarrow^*$ \\
\hline
\end{tabular}
}
\vspace{-5pt}
\end{table*}

\subsection{Overall Performance}
We present the performance of CoLD PRM, along with various baselines and base models, across two evaluation datasets: MATH500~\citep{hendrycks2024measuring} and GSM-Plus~\citep{li2024gsm}. The main results are provided in Tables~\ref{tab:baseline performance} and \ref{tab:basemodel_llama3}, and the key findings are outlined as follows:

\begin{itemize}[leftmargin=10pt]
\vspace{-5pt}
    \item \textbf{CoLD PRM achieves state-of-the-art accuracy while favoring significantly shorter steps, effectively mitigating length bias.} This effect is particularly pronounced on MATH500, where trajectories are more complex and prone to verbosity. By correcting for length-induced distortions, CoLD PRM demonstrate its robustness and superior evaluative performance in challenging scenarios.
\vspace{-5pt}
    \item As shown in Table~\ref{tab:baseline performance}, \textbf{while these baselines effectively address reward bias in RL, they struggle to generalize to length bias in PRMs}, yielding suboptimal accuracy and verbose responses. In contrast, our CoLD framework significantly improves both predictive accuracy and response conciseness.
\vspace{-5pt}
    \item Table~\ref{tab:basemodel_llama3} demonstrates that even \textbf{without joint training, combining a bias estimator with length-penalty correction substantially improves performance across various base models}. This highlights the effectiveness of these components and suggests that our framework can be flexibly applied to existing, pre-trained PRMs. In settings with limited computational resources, this provides a practical and efficient way to improve reward quality without retraining the entire model.
\vspace{-5pt}
    \item It is worth noting that \textbf{shorter length does not inherently equate to higher accuracy}. For instance, Math-Shepherd-PRM selects shorter reasoning steps yet exhibits lower accuracy. A model that aggressively favors brevity may overlook essential reasoning steps. The goal of CoLD is not to shorten responses indiscriminately, but rather to reduce unnecessary verbosity while preserving or even improving accuracy.
    \vspace{-5pt}
\end{itemize}

\begin{table*}[t]

\centering    

\vspace{-5pt}

\caption{Ablation study of CoLD PRM variants using Best-of-16 search. Solutions are sampled from Llama-3-70B-Instruct. We evaluate the contribution of each module through component-wise ablations. Additional results are available in Table~\ref{tab:ablation performance on muggle} and Table~\ref{tab:ablation performance on metamath}.
}
\vspace{-5pt}

\label{tab:ablation performance}
\resizebox{\textwidth}{!}{
\renewcommand\arraystretch{1.2}
\begin{tabular}{cccccccccc}

\hline

\multicolumn{3}{c}{Components} &  \multicolumn{2}{c}{MATH500} & \multicolumn{2}{c}{GSM-Plus} & \multicolumn{2}{c}{Avg}\\ 

 \cmidrule(r){4-5} \cmidrule(r){6-7}  \cmidrule(r){8-9}  
Joint Train &Bias Estimator& Penalty& ArithACC(\%) & Length & ArithACC(\%) & Length& ArithACC(\%) & Length  \\ 
   \hline

\checkmark &\checkmark&\checkmark& \textbf{49.2} & \textbf{313.2} &\textbf{73.8} & \textbf{202.5} & \textbf{61.5} & \textbf{257.9}\\

\checkmark&\checkmark&$\times$ &\underline{48.2} &494.9 &\underline{73.2}&\underline{237.4}&\underline{60.7}&366.2 \\
$\times$& \checkmark&\checkmark&48.0 &\underline{370.4} &72.1 & 258.6&60.1&\underline{314.5}\\
$\times$&\checkmark&$\times$& 44.8&525.0 & 71.5 & 303.6&58.2& 414.3 \\
$\times$&$\times$&\checkmark& 43.2&495.1& 71.5 & 300.1&58.2 & 397.6 \\

$\times$&$\times$&$\times$& 44.8 & 555.7&  71.0 & 324.6& 57.9 &440.2 \\

\hline   
\end{tabular}
}

\vspace{-5pt}
\end{table*}

\subsection{Ablation Study}

To understand the contribution of each component in our CoLD PRM framework, we conduct a comprehensive ablation study by selectively removing the Joint Training, Bias Estimator, and Length Penalty variants. The performance of these variants is presented in Table~\ref{tab:ablation performance}, from which we can draw the following observations:

\begin{itemize}[leftmargin=10pt]
\vspace{-5pt}

    \item \textbf{Removing Joint Training leads to a noticeable increase in the average solution length across both datasets}, indicating that without joint optimization, the reward model tends to favor longer reasoning trajectories. This suggests that joint training helps align the reward signal with downstream policy behavior, implicitly regularizing verbosity.
    \vspace{-3pt}

    \item \textbf{Excluding the Bias Estimator similarly leads to longer solutions, while the accuracy remains relatively stable.} This points to the Bias Estimator’s role in explicitly identifying and correcting for length-related confounding effects in the reward scores, encouraging the model to recognize correctness independent of surface-level length cues.

    \vspace{-3pt}

    \item \textbf{Using the Length Penalty alone reduces length but fails to improve correctness, and can even lead to a degradation in overall accuracy.} This suggests that a naive penalty on length, when not informed by bias estimation, may remove beneficial reasoning steps along with redundant ones, highlighting the risk of applying heuristic penalties without semantic guidance.
\vspace{-5pt}
\end{itemize}

\begin{wrapfigure}{R}{0.5\textwidth}
   \vspace{-15pt} 
  \begin{subfigure}[t]{0.48\linewidth}
    \centering
    \includegraphics[width=\linewidth]{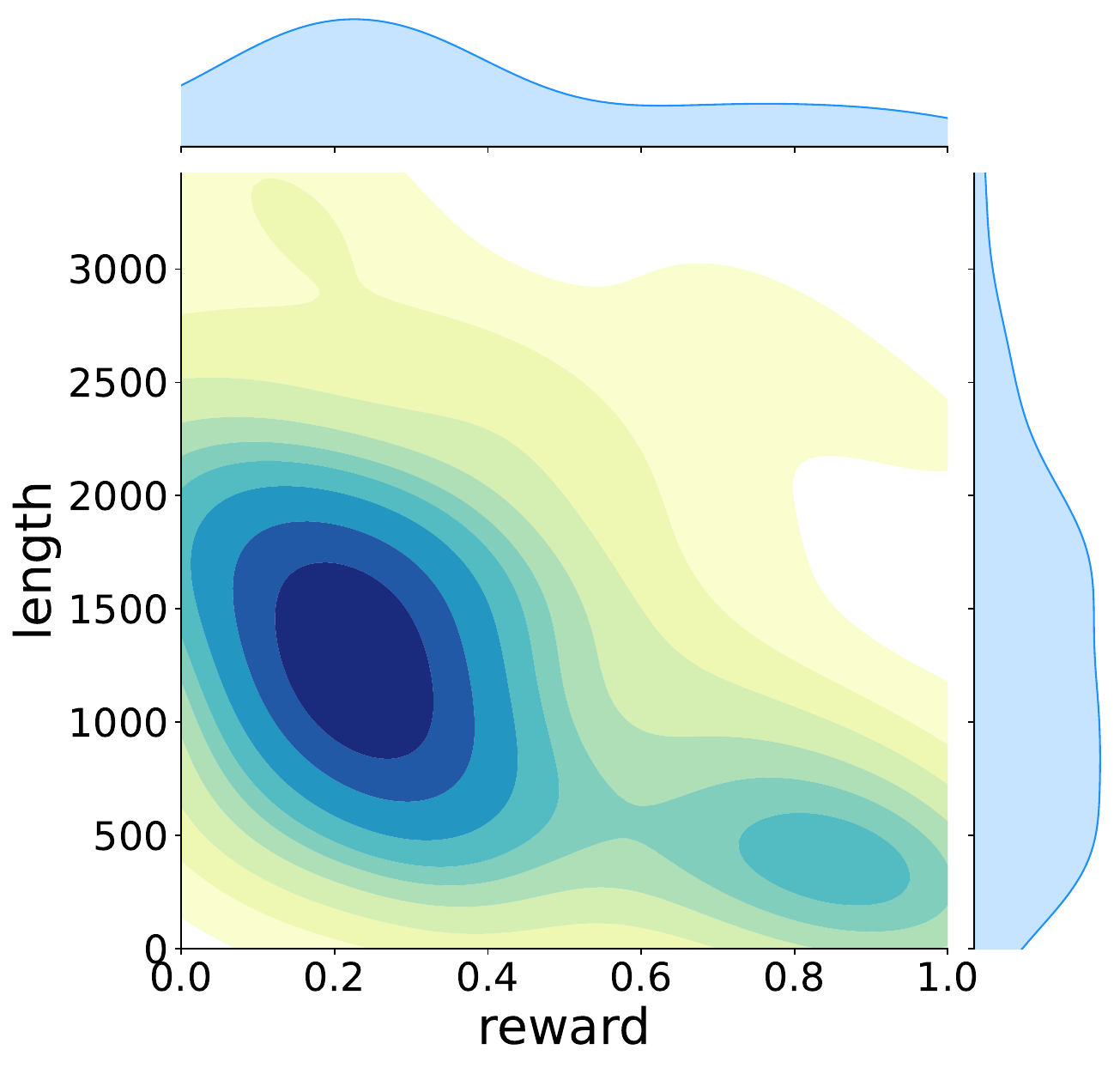}
    \caption{Original Version}
  \end{subfigure}
  \hfill
  \begin{subfigure}[t]{0.48\linewidth}
    \centering
    \includegraphics[width=\linewidth]{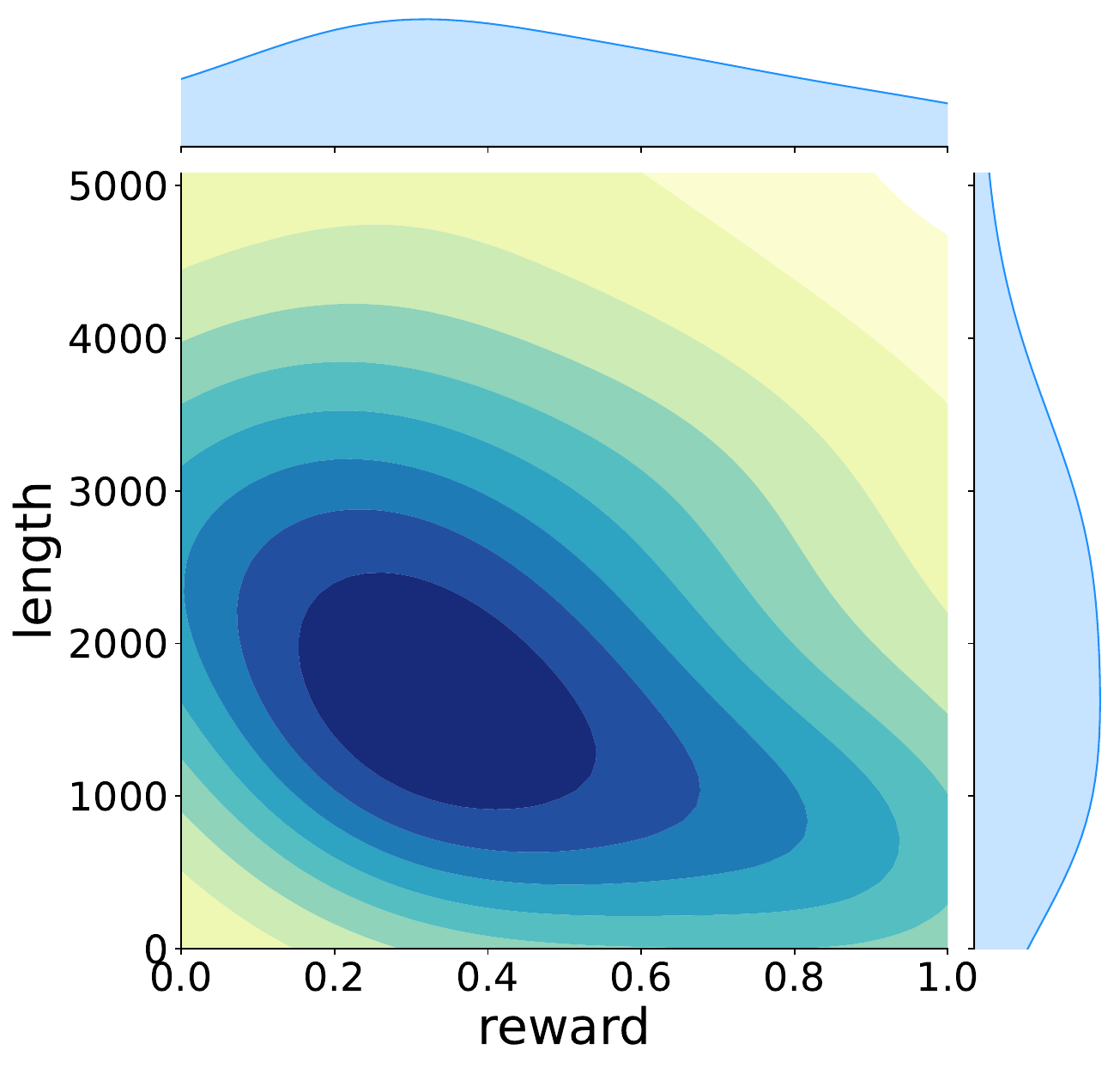}
    \caption{Extend Version}
  \end{subfigure}
  \caption{The joint distribution of rewards and step lengths after debiasing on both original and length-augmented steps.}
  \label{fig:debias-reward-length}
  \vspace{-20pt}
\end{wrapfigure}

\subsection{Debias Visualization}
To better understand the effect of our debiasing framework, we visualize the reward distributions after applying CoLD. As discussed in the introduction and shown in Figure~\ref{fig:reward-length}, the reward model favors longer reasoning steps even when they are semantically equivalent to shorter ones.

After applying CoLD, \textbf{we observe a marked improvement in the alignment between the rewards assigned to original and extended steps.} As shown in Figure~\ref{fig:debias-reward-length}, the distributions of scores become substantially more consistent, indicating that the model no longer favors more verbose steps. This suggests that CoLD PRM successfully removes the spurious correlation between step length and reward.

Notably, the debiased model places greater emphasis on the semantic correctness and logical validity of each step, rather than its superficial length. This demonstrates that our method not only improves quantitative performance but also corrects length bias in a principled manner.

%


\subsection{Downstream RL application}
To demonstrate CoLD’s effectiveness on downstream tasks, we apply CoLD in a reinforcement learning setting. Specifically, our experiments are conducted within the ReasonFlux-PRM framework~\citep{zou2025reasonflux}. We use Qwen2.5-Math-1.5B-Instruct as the policy model and adopt GRPO (Group Relative Policy Optimization)~\citep{shao2024deepseekmath} as the RL optimization algorithm. CoLD is integrated into the ReasonFlux-PRM and compared against its vanilla counterpart. For evaluation, we use MATH500~\citep{lightman2023let} as the test benchmark. 



\begin{figure}[htbp]
\vspace{-10pt}
  \centering
  \begin{minipage}{0.48\textwidth}
    \centering
    \includegraphics[width=\linewidth]{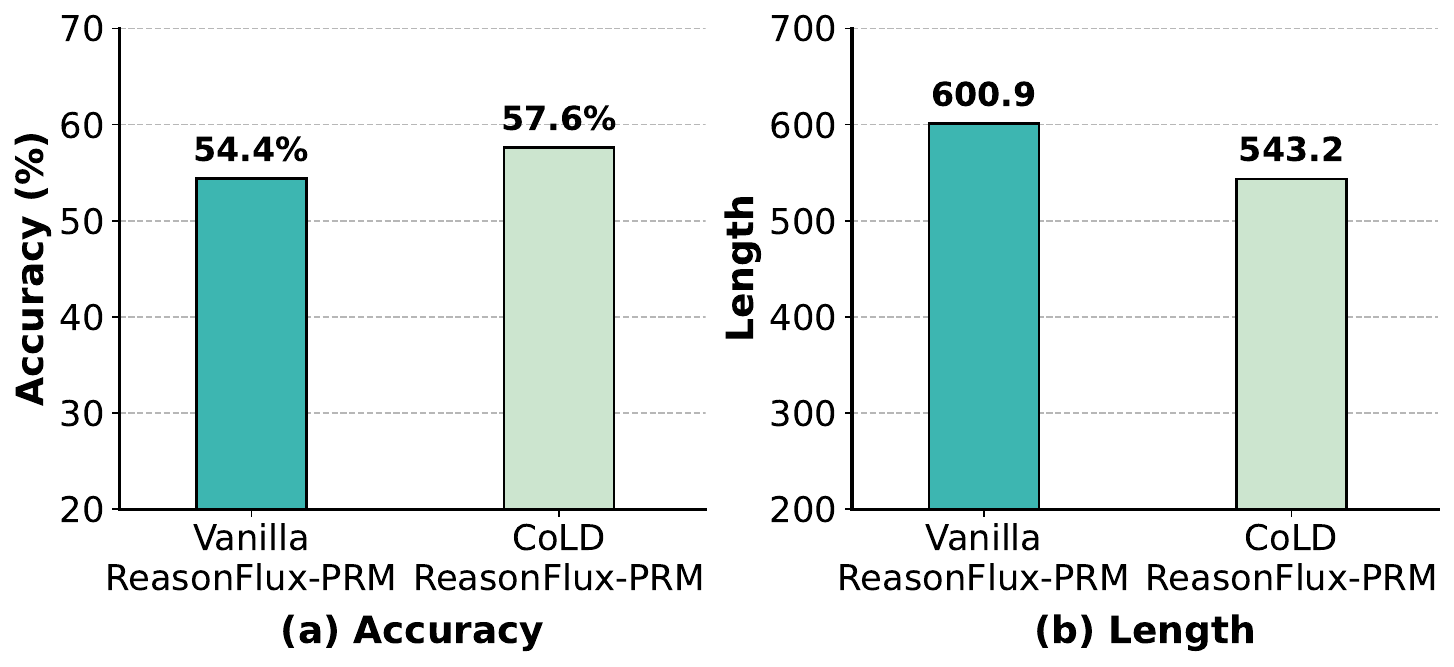}
    \vspace{-10pt}
    \caption{Performance of PRMs as reward signals in policy optimization.}
    \label{fig:rl}
  \end{minipage}
  \hfill 
  \begin{minipage}{0.48\textwidth}
    \centering
    \includegraphics[width=\linewidth]{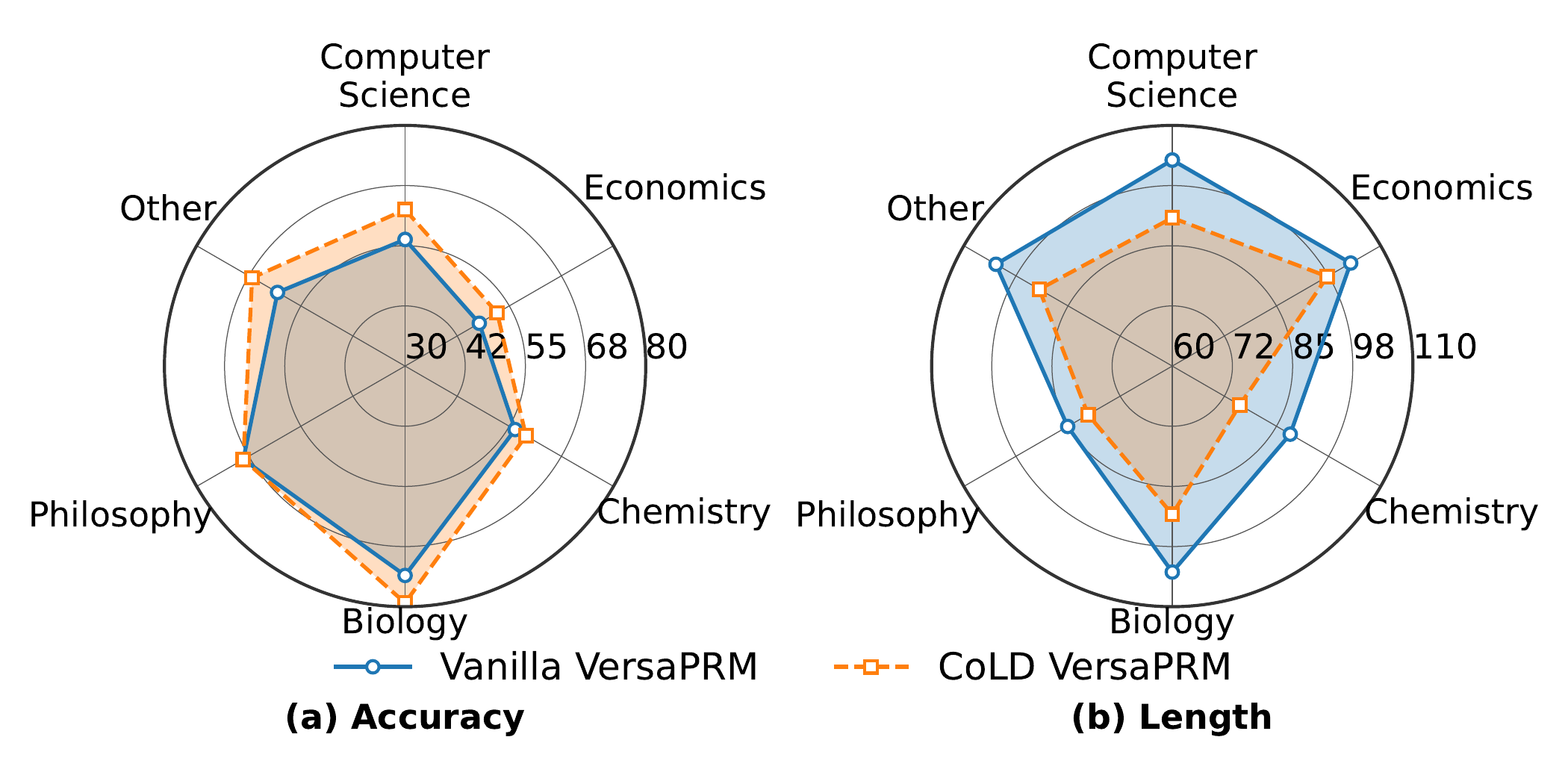}
    \vspace{-20pt}
    \caption{Generalization performance of CoLD PRM on cross-domain tasks.}
    \label{fig:cross_domain}
  \end{minipage}
  \vspace{-10pt}
\end{figure}


As shown in the Figure~\ref{fig:rl}, incorporating CoLD not only improves the accuracy of the policy model in problem solving, but also reduces the output length. This indicates that CoLD effectively mitigates the length bias exhibited by the PRM, \textbf{providing the model with more precise and meaningful reward signals during downstream reinforcement learning}.

\subsection{Cross-domain Generalization Study}
To evaluate whether CoLD can generalize to domains beyond its original scope, we integrate it into VersaPRM~\citep{zeng2025versaprm}, a process reward model designed to operate across heterogeneous domains. We assess its performance on the MMLU-Pro~\citep{wang2024mmlu} benchmark, a robust and challenging massive multi-task understanding dataset created to rigorously test the capabilities of large language models. CoLD is applied within VersaPRM across several representative domains, including Computer Science, Economics, Chemistry, Biology, Philosophy, and Other.

Based on the results shown in the Figure~\ref{fig:cross_domain}, when applying CoLD on top of VersaPRM, we observe consistent improvements in both accuracy and the reduction of response length across these heterogeneous domains. This demonstrates that the debiasing principle behind CoLD does not rely on domain-specific structures found in mathematical reasoning, but instead \textbf{addresses a general phenomenon of superficial-feature bias in step-level reward modeling}.



\subsection{Additional Experiments}

\begin{itemize} [leftmargin=10pt]
    \vspace{-3pt}
    \item \textbf{Scaling Width Study} In Appendix~\ref{app: width}, We evaluate CoLD across varying scaling width($N \in \{2, 4, 8, 16\}$), demonstrating that CoLD consistently outperforms vanilla PRMs.
\vspace{-1pt}
    \item \textbf{Hyperparameter Study} A detailed analysis of hyperparameters $c$, $\lambda_b$, and $\alpha$ is provided in Appendix~\ref{app:hyper}, highlighting CoLD’s inherent parameter robustness.
\vspace{-1pt}
    \item \textbf{Other Benchmark} In Appendix~\ref{app: Other Benchmark}, we evaluate CoLD on the representative ProcessBench benchmark. The results demonstrate that CoLD successfully mitigates bias while preserving the PRM's core evaluative capabilities.
\vspace{-1pt}
    \item \textbf{Evaluation Using Stronger Policy Model} Appendix~\ref{app:stronger model} provides additional evaluations using Qwen3-Next-80B-A3B-Instruct~\cite{yang2025qwen3} and DeepSeek-v3.1~\cite{liu2024deepseek}, where CoLD exhibits robust performance across these advanced model.
\vspace{-1pt}
    \item \textbf{Bias Estimator Interpretability} We conduct interpretability analyses focusing on its sensitivity to input length versus semantic content in Appendix~\ref{app:Interpretability}.
    \vspace{-3pt}
\end{itemize}

\section{Related Works}
\subsection{Process Reward Models}

Process Reward Models (PRMs) enable fine-grained step-level supervision for model reasoning, addressing traditional Outcome Reward Models (ORMs)~\citep{gsm8k} limitation of only scoring final outputs. This design mitigates "spurious correctness" and boosts stability in complex tasks. A key challenge for PRM training is high annotation costs: PRM800K~\citep{lightman2023let} was the first human-annotated dataset, while later works ~\citep{wang2023math,luo2024improve,zhang2025lessons} used LLMs and Monte Carlo (MC) methods to reduce costs. PRMs have advanced for both RL optimization~\citep{rizvi2025spare,chen2025discriminative,wang2024openropensourceframework,cheng2025stop,zhang2025process,feng2025prm} and inference time scaling~\citep{ma2023let,xie2025outcomes,zhao2025genprm,setlur2024rewarding,hu2025coarse,hu2025prm}. Despite these advances, few PRM-related works have focused on the length bias issue inherent in PRMs—and this is precisely the problem addressed by our proposed CoLD PRM.

\subsection{Causal Debias}
Causal debias has emerged as a fundamental strategy across diverse machine learning domains~\citep{zhu2024m,yu2025causal,sun2025causal}. For example,~\citet{zhang2024causal} mitigates bias in multi-hop fact verification by applying front-door adjustment to reasoning paths and estimating causal effects via random walks. ~\citet{chisca2024prompting} reduce LLM bias by leveraging causal paths to design prompts that emphasize factual knowledge over stereotypes. ~\citet{zhou-etal-2023-causal} propose a fine-tuning framework that integrates causal intervention and invariant risk minimization to suppress reliance on non-causal, bias-inducing factors. ~\citet{liu2025target} introduces a counterfactual-enhanced framework that debiases multimodal sentiment classification through adaptive contrastive learning. ~\citet{liu2024rrm} propose a causal framework with data augmentation to filter out irrelevant artifacts. ~\citet{zhan2023debiasing} presents a counterfactual training strategy for Med-VQA, removing spurious linguistic correlations via causal intervention. ~\citet{farzam2024causal} analyzes how differential privacy can bias causal estimates and proposes robust regression techniques to correct such distortions.  
While these methods effectively address domain-specific confounders, none directly tackle the unique length bias~\citep{park2024disentangling,dubois2024length} in process reward models (PRMs). Our proposed CoLD PRM fills this gap by disentangling semantic correctness from spurious correlations with output length.

\section{Conclusion}
In this work, we identify and systematically study length bias in process reward models (PRMs). To tackle this issue, We introduce CoLD, a Counterfactually-Guided Length Debiasing framework. Specifically, CoLD incorporates a length penalty, a bias estimator, and joint training to mitigate the undue influence of response length on reward scores. Extensive experiments show that CoLD improves solution-selection accuracy and RL performance by mitigating length bias, establishing a generalizable and robust strategy for process reward modeling.

\bibliography{example_paper}
\bibliographystyle{ACM-Reference-Format}

\appendix


\section{Causal Counterfactual Analysis}
\label{sec:Causal Counterfactual Analysis}
In the causal graph shown in Figure~\ref{fig:causal graph}, variables influence one another. For example, both correctness $C$ and step length $L$ affect the prediction $P$. Therefore, the value of $P$ can be determined by its ancestor nodes, mathematically expressed as:
\begin{equation}
    P_{c,n,l}=P(C=c,N=n,L=\ell).
\end{equation}
where $P(\cdot)$ denotes the value function associated with $P$.

To rigorously quantify the effect of these variables on the model prediction, we adopt counterfactual methods. The core idea of counterfactual analysis is to evaluate the outcome under hypothetical scenarios by selectively altering certain variables. For example, we can set $L$ to a fixed reference value $\ell^*$, which represents a counterfactual intervention that removes its actual effect from the system. Since $\ell^*$ is held constant, it acts as a reference condition with a consistent influence on downstream variables. Likewise, we counterfactually set $C$ to $c^*$ to isolate the combined effect of $C$ and $L$ on $P$. Under this setting, the total effect on $P$ can be formulated as:
\begin{equation}
    E_{total}=P_{c,n,\ell}-P_{c^*,n^*,\ell^*}
\end{equation}
Here, $C$ is also replaced by $c^*$ because $c^*$ also has an influence on $P$. $c^*$ represents the counterfactual value of $P$.

Moreover, based on the structure of the causal graph, we can decompose this total effect into two distinct components: (1) the effect of the correctness variable on the PRM prediction, i.e., the path $C \rightarrow P$, and (2) the effect of length, i.e., $L \rightarrow P$. To isolate the effect of length, we hold $C$ at its counterfactual value $c^*$ while allowing $L$ to vary, which gives:
\begin{equation}
    E_{L \rightarrow P}=P_{c^*,n^*,\ell}-P_{c^*,n^*,\ell^*}
\end{equation}
where $P_{c^*,L}$ represents the prediction when the observed length $L$ is retained, while the correctness signal $C$ has been counterfactually removed as shown in Figure~\ref{fig:causal graph}(b).

We perform this counterfactual removal of $C$ because we aim to measure only the contribution of $L$ to the prediction, independent of $C$. This process cannot be directly computed from observational data and thus relies on causal reasoning, making it a canonical example of counterfactual causal inference.

Once $E_{total}$ and $E_{L \rightarrow P}$ are calculated, $E_{C,N \rightarrow P}$ can be obtained by subtracting the former from the latter, as illustrated in Figure~\ref{fig:causal graph}:
\begin{equation}
\begin{split}
    E_{C,N \rightarrow P}&=E_{total}-E_{L \rightarrow P}\\
                    &=P_{c,n,\ell}-P_{c^*,n^*,\ell^*}-(P_{c^*,n^*,\ell}-P_{c^*,n^*,\ell^*})\\
                    &=P_{c,n,\ell}-P_{c^*,n^*,l}\\
                    &=P(C=c,N=n,L=\ell)-P(C=c^*,N=n^*,L=\ell)
\end{split}
\end{equation}

This expression captures the isolated effect of the correctness signal $C$ on the model prediction $P$, with length $L$ held constant. As discussed in Section~\ref{sec:method}, the value $P(C = c, L = \ell)$ can be estimated using the composed reward $r_\theta(x)\sigma(b_\phi(x + \mathcal{N}))$, where $r_\theta(x)$ is the PRM score and $b_\phi(x + \mathcal{N})$ is the bias estimator output. On the other hand, $P(C = c^*, L = \ell)$ can be approximated by both $\sigma(b_\phi(x + \mathcal{N}))$ and $\alpha \ell(x)$.

Importantly, the training labels themselves inherently contain verbosity bias, and it is practically impossible to directly obtain truly unbiased labels in real-world scenarios. This is precisely why our framework is designed in this way. During training, we use the CE loss to align the composed, biased term $r_\theta(x)\sigma(b_\phi(x+\mathcal{N}))$ with the biased labels, enabling the joint model to properly fit the observational data distribution. During inference, however, we apply Eq.~3 to subtract the counterfactual bias term from the prediction. In this way, our method remains faithful to the causal interpretation: we fit the biased factual distribution during training, and then perform counterfactual intervention at inference to recover the true, debiased semantic reward.

\begin{algorithm*}[t]
\caption{Joint Training of PRM and Bias Estimator}
\label{alg:joint-training}
\begin{algorithmic}[1]
\REQUIRE Training dataset $\mathcal{D} = \{(x_i, y_i)\}_{i=1}^N$, learning rates $\eta_r$, $\eta_b$, hyperparameters $\lambda_r$, $\lambda_b$, bias correction factor $c$
\STATE Initialize PRM $r_\theta(\cdot)$ and Bias Estimator $b_\phi(\cdot)$
\WHILE{not converged}
    \STATE Sample mini-batch $\{(x_i, y_i)\}_{i=1}^B \sim \mathcal{D}$
    \FOR{each $x_i$ in batch}
        \STATE Inject noise: $\tilde{x}_i = x_i + \mathcal{N}$
        \STATE Compute $\hat{r}_i = r_\theta(x_i) \cdot \sigma\bigl(b_\phi(\tilde{x}_i)\bigr)$
    \ENDFOR
    \STATE Compute cross-entropy loss:
    \[
        \mathcal{L}_{\text{CE}} = -\frac{1}{B} \sum_{i=1}^B \left[ y_i \log \sigma(\hat{r}_i) + (1 - y_i) \log (1 - \sigma(\hat{r}_i)) \right]
    \]
    \STATE Compute Pearson correlations $\rho_r = \mathrm{Corr}(r_\theta(x), \ell(x))$, $\rho_b = \mathrm{Corr}(b_\phi(x), \ell(x))$
    \STATE Compute module-specific losses:
    \[
        \mathcal{L}_{\text{PRM}} = \lambda_r \cdot \rho_r^2, \quad
        \mathcal{L}_{\text{Bias}} =- \lambda_b \cdot \rho_b^2
    \]
    \STATE Compute final losses:
    \[
        \mathcal{L}_{\text{Final}} = \mathcal{L}_{\text{CE}} +  \mathcal{L}_{\text{PRM}}+ \mathcal{L}_{\text{Bias}}
    \]
    \STATE Update PRM parameters $\theta$: $\theta \leftarrow \theta - \eta_r \nabla_\theta (\mathcal{L}_{\text{CE}}+\mathcal{L}_{\text{PRM}})$
    \STATE Update Bias Estimator parameters $\phi$: $\phi \leftarrow \phi - \eta_b \nabla_\phi (\mathcal{L}_{\text{CE}}-\mathcal{L}_{\text{Bias}})$
\ENDWHILE
\end{algorithmic}
\end{algorithm*}

\section{Algorithm Description}
\label{appendix:alg-desc}

Algorithm~\ref{alg:joint-training} summarizes the joint training procedure for the Process Reward Model (PRM) and the Bias Estimator. The training aims to disentangle semantic correctness from length-related bias in the reward predictions. 

At each iteration, a mini-batch of training samples is drawn from the dataset. For each sample, noise is injected into the input features before feeding them to the Bias Estimator to prevent it from capturing semantic information. The composed reward $\hat{r}(x)$ is then computed as the element-wise product of the PRM output and the sigmoid-activated Bias Estimator output.

The binary cross-entropy loss supervises the composed reward to align with ground-truth correctness labels. To encourage the PRM to focus on correctness rather than spurious length correlations, a regularization term penalizing the squared Pearson correlation between PRM predictions and step length is added. Conversely, the Bias Estimator is encouraged to maximize its correlation with step length to model the bias effectively.

Separate gradient updates are applied to the PRM and Bias Estimator parameters using their respective loss functions, allowing each module to specialize in modeling semantic correctness and length bias, respectively.

At inference time, the debiased reward is computed by correcting the PRM output with the estimated bias from the Bias Estimator, scaled by a hyperparameter controlling the bias removal magnitude.

\section{Experiment Details}
\label{app:Experiment Details}

\subsection{Implementation Details}
\label{app:Implementation Details}
\paragraph{Train Dataset}
We use both PRM800K and Math-Shepherd as training datasets. The first, PRM800K~\citep{lightman2023let}, consists of 800,000 step-level correctness labels derived from the MATH dataset via extensive human annotation, offering a high-fidelity but annotation-expensive training resource. The second, Math-Shepherd~\citep{wang2024math}, comprises 400,000 automatically generated labels across both MATH and GSM8K problems. It provides scalable supervision without human involvement, allowing for cost-effective training at scale.
\paragraph{Evaluation Dataset}
The evaluation datasets employed in our study are curated by~\citet{li2024process}, drawing from the MATH500~\citep{hendrycks2024measuring} and GSM-Plus~\citep{li2024gsm} datasets. The collected trajectories are sampled from three mathematical solvers of varying model scales: MetaMath-Mistral-7B~\citep{yu2023metamath}, MuggleMath-13B~\citep{li2023query}, and Llama3-70B-Instruct~\citep{llama3modelcard}. In addition, we construct a semi-synthetic extension of this dataset to facilitate controlled evaluation of length-related biases. Specifically, for each original solution trajectory, we generate semantically equivalent but longer variants using two strategies: (1) duplicating individual steps to create trivially lengthened versions, and (2) prompting DeepSeek~\citep{liu2024deepseek} to rewrite each step with increased verbosity while preserving its semantic meaning and logical validity. These two transformation methods reflect two common patterns in model-generated outputs—verbatim repetition and verbose paraphrasing with limited substantive contribution. An example illustrating the two expansion methods can be found in Figure~\ref{fig: example}.

To further verify that LLM-based sentence expansion does not alter the original meaning (i.e., correctness), we randomly sampled 300 pairs of expanded and original sentences for manual evaluation. The results show a pass rate of 95.3\%, indicating that the imposed constraints on sentence rewriting are effective and largely preserve semantic fidelity.

Based on the above, we construct a test set consisting of 16 solutions per question: 8 original trajectories and 8 length-augmented variants that retain the same underlying semantics. This design enables fine-grained analysis of model behavior under superficial variations in step length.

\paragraph{Hyperparameter and Training Settings}
For joint training, we adopt a batch size of 128, with initial learning rates of \(1 \times 10^{-4}\) for the PRM and \(3 \times 10^{-4}\) for the bias estimator. The loss weights are set as \(\lambda_r = 0.1\) and \(\lambda_b = 0.5\) throughout training.

For experiments that train only the bias estimator, we use a batch size of 64. When training on the Math-Shepherd PRM, the initial learning rate is set to \(2 \times 10^{-3}\), with \(\lambda_{\text{corr}} = 0.3\). When using our own PRM (trained on PRM800K), we set the learning rate to \(2 \times 10^{-3}\), also with \(\lambda_{\text{corr}} = 0.3\).

We selected Qwen-2.5-Math-7B-instruct \citep{yang2024qwen2} as the foundational large language model (LLM) for our PRM, and Qwen-2.5-0.5B-instruct as the base model for the bias estimator. For joint training, four NVIDIA A100 GPUs were utilized, with the training process taking approximately 5 hours to complete. While solely training the bias estimator, only one RTX 4090 GPU was required, and this training procedure lasted around 3 hours. To enhance training resource efficiency, we employed Parameter-Efficient Fine-tuning techniques LoRA. The LoRA configuration was set with a rank of 8, an alpha value of 32, and dropout set to 0.1.

\paragraph{Debiasing Methods in RLHF}
\begin{itemize}
    \item \textbf{Length penalty\citep{singhal2023long}:} The method adds a linear proportional penalty term to the original reward
    \item \textbf{Loose Lips Sink Ships\citep{shen2023loose}:}The method applies a Product-of-Experts framework that disentangles reward modeling from sequence-length effects by combining a semantic expert focused on human intent with a perturbed bias expert specialized in capturing length bias.
    \item \textbf{Adaptive Length Bias Mitigation\citep{bu2025beyond}:}The method introduces Adaptive Length Bias Mitigation (ALBM), which disentangles length bias from the original reward and adaptively recombines the length and quality rewards based on the characteristics of each query.
    \item \textbf{Uniform Average\citep{huang2024post}:} The method uses the local average reward to provide an estimation for the bias term, which can be removed, thereby approximating the true reward.
    \item \textbf{Locally Weighted Regression(LWR)\citep{huang2024post}:} The method assigns weights to nearby data points, giving higher importance to those closer to the target. This ensures that proximate points significantly influence the regression. LWR then applies weighted linear regression to model the local behavior of the target function, effectively approximating the weighted average within the local context.
\end{itemize}

\vspace{2em}

\paragraph{Basemodels}
\begin{itemize}
    \item \textbf{Math-Shepherd-PRM\citep{wang2023math}:} generates process labels by estimating the empirical probability that a step leads to the correct answer and trains a Process Reward Model (PRM) on their published dataset.
    \item \textbf{Base-PRM:} trained by ourselves using the human-annotated PRM800K dataset, based on Qwen-2.5-Math-7B-instruct \citep{yang2024qwen2}.
    \item \textbf{Qwen2.5-Math-PRM\citep{zhang2025lessons}:} adopts a two-phase data construction (data expansion, data filtering) and specific training. In expansion, it uses MC estimation with hard labels (a response is negative only if no 8 completions get the correct answer). In filtering, it uses Qwen2.5-Instruct-72B as LLM-as-a-judge to verify reasoning step-by-step, and applies consensus filtering to remove instances with mismatched LLM-annotated and MC-estimated process labels for quality. For training, it uses cross-entropy loss on end-of-step tokens for binary classification.
    \item \textbf{EurusPRM-Stage1\citep{cui2025process}:} is trained via Implicit PRM\citep{yuan2024free}, a framework that secures free process rewards without incurring additional costs—requiring only the simple training of an ORM (Outcome Reward Model) on more affordable response-level labels. During inference, implicit process rewards are generated by performing a forward pass and computing the log-likelihood ratio at each step.
    \item \textbf{ReasonFlux-PRM~\citep{zou2025reasonflux}:}a trajectory-aware preference model that integrates step-level and trajectory-level supervision to provide fine-grained reward signals aligned with structured reasoning traces.
    \item \textbf{VersaPRM~\citep{zeng2025versaprm}:}a multi-domain PRM trained on synthetic reasoning data produced through novel data generation and annotation pipeline.
\end{itemize}
\subsection{Example of Semi-synthetic Solution}
\label{app:Example}
\begin{figure}[t]
  \centering
  \includegraphics[width=\textwidth]{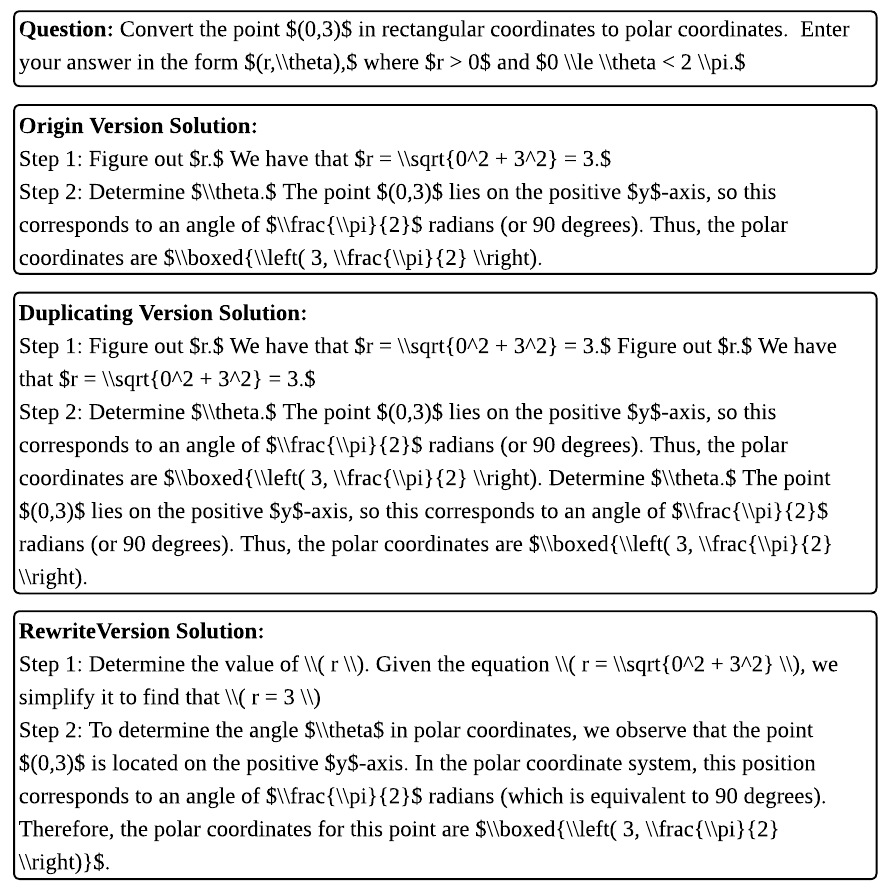}
  \caption{An example of the original and extended solutions}
  \label{fig: example}

\end{figure}

\begin{figure}[t]
  \centering
  \includegraphics[width=\textwidth]{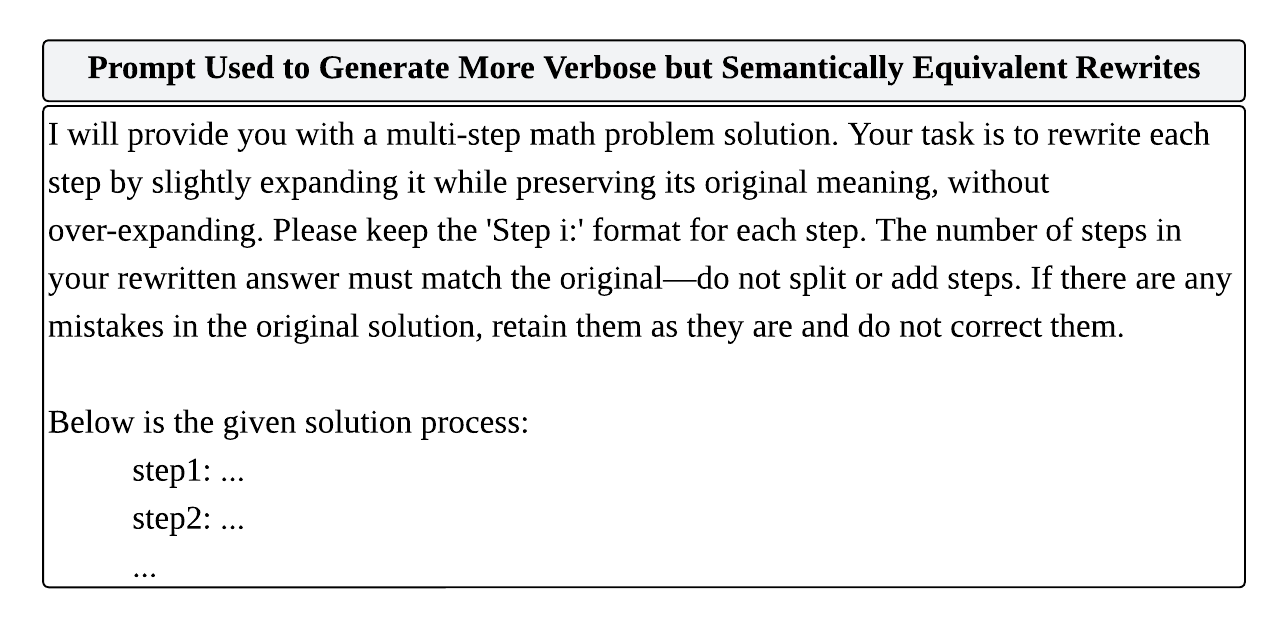}
  \caption{Prompt Used to Generate More Verbose but Semantically Equivalent Rewrites by DeepSeek}
  \label{fig: prompt}

\end{figure}

We generate extended variants either by duplicating the original step or by prompting DeepSeek to produce more verbose yet semantically equivalent rewrites, as illustrated in Figure~\ref{fig: example}. The prompt used for this process is provided in Figure~\ref{fig: prompt}.

\subsection{RL Experiment Settings}
We conducted a reinforcement learning experiment using the Group Relative Policy Optimization (GRPO)~\citep{shao2024deepseekmath} algorithm via the verl framework. Specifically, we fine-tuned the Qwen2.5-1.5B-Instruct~\citep{yang2024qwen2} model. The training process utilized the dapo\_500steps dataset with a maximum response length of 16,384 tokens to support long-form reasoning, while the GRPO group size was set to $G=6$ per prompt. For computational efficiency, the experiment was executed on a single node with 4 GPUs, employing a learning rate of $1 \times 10^{-6}$ and a KL coefficient of $0.001$. We optimized memory usage by enabling gradient checkpointing, chunked prefill, and CPU parameter offloading for the reference model, while the rollout engine utilized vLLM with a tensor parallel size of 2 to maintain high throughput during the generation phase.


\section{Supplementary Experiments}
In this section, we present additional results across different policy models, as shown in Table~\ref{tab:basemodel_metamath}, Table~\ref{tab:basemodel_muggle}, Table~\ref{tab:ablation performance on muggle} and Table~\ref{tab:ablation performance on metamath}.

\begin{table*}[t]
\centering
\caption{Generalization of CoLD using the MetaMath-Mistral-7B policy model. Performance is evaluated using Best-of-16 accuracy. The $\uparrow$ indicates an increase in accuracy or length, while the $\downarrow$ denotes a decrease. A paired t-test was conducted to verify results. Statistically significant improvements over the baseline ($p < 0.05$) are denoted with an asterisk (*).}
\label{tab:basemodel_metamath}
\resizebox{\textwidth}{!}{
\renewcommand\arraystretch{1.2}
\begin{tabular}{lcccccc}
\hline
\multirow{2}{*}{\textbf{PRM Model}} & \multicolumn{2}{c}{\textbf{MATH500}} & \multicolumn{2}{c}{\textbf{GSM-Plus}} & \multicolumn{2}{c}{\textbf{Avg}}  \\ 
\cmidrule(r){2-3} \cmidrule(r){4-5} \cmidrule(r){6-7}
 & \textbf{ArithACC(\%)} & \textbf{Length} & \textbf{ArithACC(\%)} & \textbf{Length}& \textbf{ArithACC(\%)} & \textbf{Length}\\ 
\hline 
Vanilla-Math-Shepherd-PRM & 32.6 & 381.3 & 59.9 & 291.8 & 46.3 & 336.6 \\
CoLD(w/o Joint)-Math-Shepherd-PRM & 32.6 & 300.1$\downarrow^*$ & 60.4$\uparrow$ & 279.9$\downarrow^*$ & 46.5$\uparrow$ & 290.0$\downarrow^*$ \\
Vanilla-Qwen2.5-Math-PRM & 36.4 &648.8 & 62.8 & 292.3 & 49.6 & 470.6 \\
CoLD(w/o Joint)-Qwen2.5-Math-PRM & 39.4$\uparrow^*$ & 338.4$\downarrow^*$ & 62.5$\downarrow$ & 235.0$\downarrow^*$ & 51.0$\uparrow$ & 286.7$\downarrow^*$ \\
Vanilla-EurusPRM-Stage1 & 43.2 &324.0 &65.4 & 323.6 & 54.3 & 357.7 \\
CoLD(w/o Joint)-EurusPRM-Stage1 & 42.0$\downarrow$ &239.7$\downarrow^*$ & 64.8$\downarrow$ & 260.8$\downarrow^*$ & 53.4$\downarrow$ & 250.3$\downarrow^*$ \\
\hline
\end{tabular}
}
\end{table*}

\begin{table*}[t]
\centering
\caption{Generalization of CoLD using the Muggle-Math-13B policy model. Performance is evaluated using Best-of-16 accuracy. The $\uparrow$ indicates an increase in accuracy or length, while the $\downarrow$ denotes a decrease. A paired t-test was conducted to verify results. Statistically significant improvements over the baseline ($p < 0.05$) are denoted with an asterisk (*).}
\label{tab:basemodel_muggle}
\resizebox{\textwidth}{!}{
\renewcommand\arraystretch{1.2}
\begin{tabular}{lcccccc}
\hline
\multirow{2}{*}{\textbf{PRM Model}} & \multicolumn{2}{c}{\textbf{MATH500}} & \multicolumn{2}{c}{\textbf{GSM-Plus}} & \multicolumn{2}{c}{\textbf{Avg}}  \\ 
\cmidrule(r){2-3} \cmidrule(r){4-5} \cmidrule(r){6-7}
 & \textbf{ArithACC(\%)} & \textbf{Length} & \textbf{ArithACC(\%)} & \textbf{Length}& \textbf{ArithACC(\%)} & \textbf{Length}\\ 
\hline 
Vanilla-Math-Shepherd-PRM & 28.6 & 332.1 & 59.2 & 279.7 & 43.9 & 305.9 \\
CoLD(w/o Joint)-Math-Shepherd-PRM & 28.8$\uparrow$ & 262.7$\downarrow^*$ & 59.1$\downarrow$ & 263.3$\downarrow^*$ & 44.0$\uparrow$ & 263.0$\downarrow^*$ \\
Vanilla-Qwen2.5-Math-PRM & 30.2 & 399.0 & 60.8 & 251.1&45.5 & 325.1 \\
CoLD(w/o Joint)-Qwen2.5-Math-PRM & 31.8$\uparrow^*$ & 297.1 $\downarrow^*$& 60.3$\downarrow$ & 222.9$\downarrow^*$ & 46.1$\uparrow$ & 260.0$\downarrow^*$ \\
Vanilla-EurusPRM-Stage1 & 34.0 &359.1 &57.6 & 300.6 & 45.8 & 328.4 \\
CoLD(w/o Joint)-EurusPRM-Stage1 & 34.6$\uparrow$ & 282.7$\downarrow^*$ & 57.9$\uparrow$ & 265.7$\downarrow^*$ & 46.3$\uparrow$ & 274.2$\downarrow^*$ \\
\hline  
\end{tabular}
}
\end{table*}

\begin{table}[t]
\centering    

\caption{We assess the performance of CoLD PRM variants using Best-of-16 search with solutions sampled from the Muggle-Math-13B model. Component-wise ablations are conducted to evaluate the contribution of each module.}

\label{tab:ablation performance on muggle}
\resizebox{1.0\textwidth}{!}{
\renewcommand\arraystretch{1.1}
\begin{tabular}{cccccccccc}
\hline
\multicolumn{3}{c}{Components} &  \multicolumn{2}{c}{MATH500} & \multicolumn{2}{c}{GSM-Plus} & \multicolumn{2}{c}{Avg}\\ 
\cmidrule(r){4-5} \cmidrule(r){6-7}  \cmidrule(r){8-9}  
Joint Train &Bias Estimator& Penalty& ArithACC(\%) & Length & ArithACC(\%) & Length& ArithACC(\%) & Length  \\ 
\hline 
\checkmark &\checkmark&\checkmark& \textbf{31.4} & \textbf{309.2} &\underline{60.3} & \underline{243.3} & \textbf{45.9} & \textbf{276.3}\\
\checkmark&\checkmark&$\times$ &\underline{30.6} &387.7 &\textbf{60.4}&273.1&\underline{45.5}&330.4 \\
$\times$& \checkmark&\checkmark&31.0 &\underline{329.0} &59.9 & 238.3&45.5&\underline{283.7}\\
$\times$&\checkmark&$\times$& 29.8&398.3 & 59.1 & 286.5&44.5& 342.4 \\
$\times$&$\times$&\checkmark& 30.0&376.1&59.3 & 280.5&44.7 & 328.3 \\
$\times$&$\times$&$\times$& 30.4 & 411.1&  59.1 & 287.9& 44.8 &349.5 \\
\hline   
\end{tabular}

}
\end{table}

\begin{table}[t]
\centering    

\caption{We assess the performance of CoLD PRM variants using Best-of-16 search with solutions sampled from the MetaMath-Mistral-7B model. Component-wise ablations are conducted to evaluate the contribution of each module.}

\label{tab:ablation performance on metamath}
\resizebox{1.0\textwidth}{!}{
\renewcommand\arraystretch{1.1}
\begin{tabular}{cccccccccc}
\hline
\multicolumn{3}{c}{Components} &  \multicolumn{2}{c}{MATH500} & \multicolumn{2}{c}{GSM-Plus} & \multicolumn{2}{c}{Avg}\\ 
\cmidrule(r){4-5} \cmidrule(r){6-7}  \cmidrule(r){8-9}  
Joint Train &Bias Estimator& Penalty& ArithACC(\%) & Length & ArithACC(\%) & Length& ArithACC(\%) & Length  \\ 
\hline 

\checkmark &\checkmark&\checkmark& 37.2 & \underline{376.3}&\textbf{61.4} & \textbf{238.6} & \textbf{49.3} & \textbf{307.5} \\

\checkmark&\checkmark&$\times$ &36.2 &552.6 &\underline{61.3}&289.7&48.8&421.2 \\

$\times$& \checkmark&\checkmark&\textbf{38.6} &\textbf{353.4}&59.8 & \underline{262.7}&\underline{49.2}&\underline{308.1} \\

$\times$&\checkmark&$\times$& \underline{38.0}&554.1 & 58.9 & 322.8&48.5& 438.4 \\

$\times$&$\times$&\checkmark& 37.6&448.8&59.1&313.9&48.4 & 381.4 \\

$\times$&$\times$&$\times$& 30.4 & 411.1& 59.4& 335.5& 44.9 & 373.3 \\

\hline   
\end{tabular}

}
\end{table}

\section{Scaling Width Study}
\label{app: width}
In this section, we analyze the impact of our CoLD method under varying numbers of best-of-N samples. 
Experiments were conducted with $N = 2, 4, 8, 16$. 

\begin{figure}[ht]
  \centering

  \includegraphics[width=\textwidth]{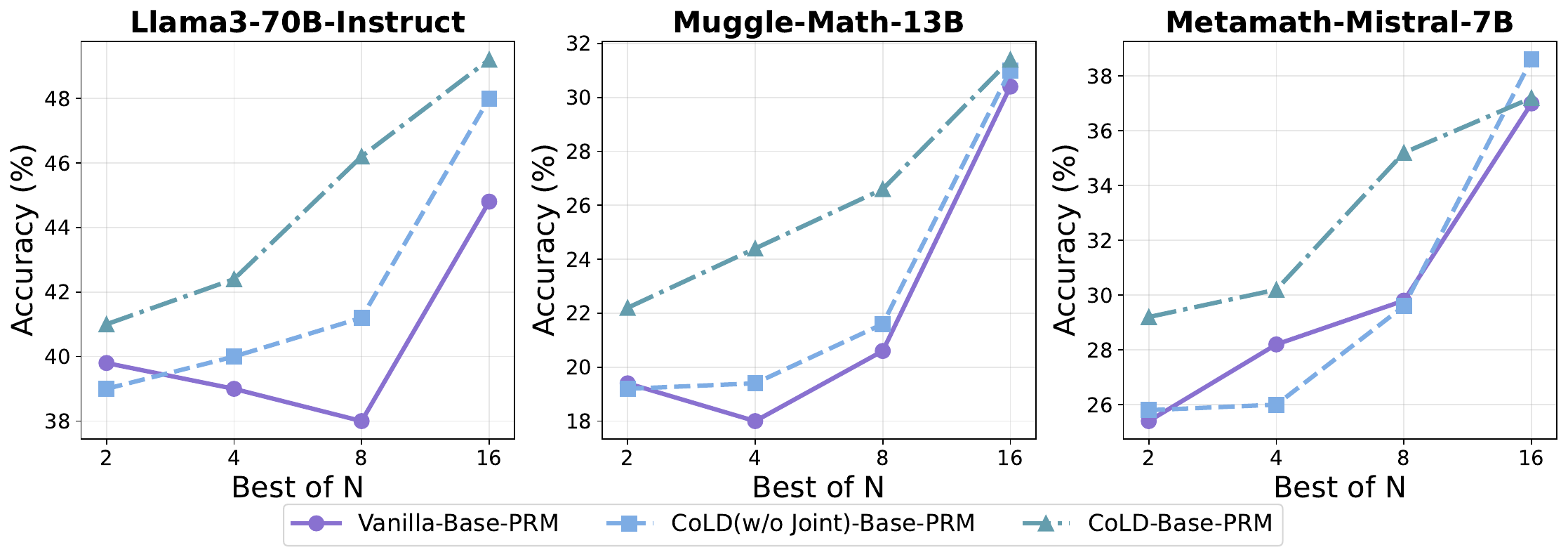}
  \caption{Performance of CoLD under different numbers of best-of-N samples on the Math dataset across different policy models.}
  \label{fig: bon}

\end{figure}

As shown in Figure~\ref{fig: bon}, our CoLD PRM consistently achieves a clear advantage across different scaling widths. For the CoLD PRM without joint training, while exhibiting slightly lower accuracy than the vanilla PRM at a few points, it outperforms the vanilla counterpart for most values of $n$. For the vanilla PRM, however, we observe that in some cases increasing the scaling width can paradoxically lead to a decline in accuracy despite the expectation that a larger $n$ should provide more candidate solutions and thus improve performance. This counterintuitive result highlights the substantial impact of length bias on the model’s ability to correctly assess solution steps.

\section{Impact of the Noise Scale $\mathcal{N}$}
\label{app:noise}
In our feature-space perturbation design, the noise scale $k$ controls the intensity of semantic corruption imposed on the underlying representation. Specifically, we inject Gaussian noise into token IDs according to the following formulation:
\begin{equation}
\tilde{x} = \text{clip}(x + k\epsilon, \, 0, \, V-1),
\end{equation}
where $\epsilon \sim \mathcal{N}(0,1)$ is truncated to the interval $[-1, 1]$, and $V$ denotes the vocabulary size. The final perturbed token IDs are obtained by rounding down, i.e., $\lfloor \tilde{x} \rfloor$. In our main experiments, we set $k=5$. To systematically evaluate the rationality of this design choice and its boundary effects on model performance, we conduct a quantitative ablation study over different values of $k$ to examine the robustness of the CoLD framework under varying noise intensities.

As shown in Table~\ref{tab:noise_scale}, when the noise injection mechanism is completely removed (i.e., $k=0$), the model’s accuracy on both benchmarks drops noticeably, while the average length of the selected solutions increases. This phenomenon suggests that, in the absence of perturbation, the bias estimator inevitably leverages its neural approximation capacity to capture and exploit deep semantic information correlated with reasoning quality in the original feature space. Such “semantic leakage” undermines the functional decomposition in the joint optimization framework, preventing the bias estimator from being effectively constrained to focus on stripping away superficial redundant features—such as step length—that are unrelated to logical correctness. As a result, its role as a dedicated bias modeling module is compromised.

In sharp contrast, when noise is introduced with $k \in \{5, 10, 15\}$, the model exhibits high stability in both prediction accuracy and length metrics. These results collectively demonstrate that while noise injection is a crucial component for effectively characterizing and modeling length bias, the overall framework remains robust at the feature-disentanglement level and is not sensitive to moderate variations in the noise scale. Therefore, in practical deployment, this mechanism can operate reliably without requiring delicate hyperparameter tuning, thereby reducing the overall cost of usage and maintenance.

\begin{table}[htbp]
\centering  

\caption{Performance of the CoLD framework under different noise scales $k$. Responses are sampled from the Llama-3-70B-Instruct model.}
\label{tab:noise_scale}
\resizebox{0.75\textwidth}{!}{
\renewcommand\arraystretch{1.2}
\begin{tabular}{ccccc}
\hline
\multirow{2}{*}{\textbf{Noise Scale ($k$)}} & \multicolumn{2}{c}{\textbf{MATH500}} & \multicolumn{2}{c}{\textbf{GSM-Plus}}  \\ 
\cmidrule(r){2-3} \cmidrule(r){4-5} 
 & \textbf{ArithACC (\%)} & \textbf{Length} & \textbf{ArithACC (\%)} & \textbf{Length}\\ 
\hline 

0 (No noise) & 47.2 & 337.9 & 71.4 & 238.2 \\
5  & 49.2 & 313.2 & 73.8 & 202.5 \\
10 & 48.9 & 308.6 & 73.1 & 200.7 \\
15 & 48.8 & 301.1 & 73.5 & 210.8 \\

\hline
\end{tabular}
}
\end{table}

\begin{figure}[t]
  \centering

  \includegraphics[width=\textwidth]{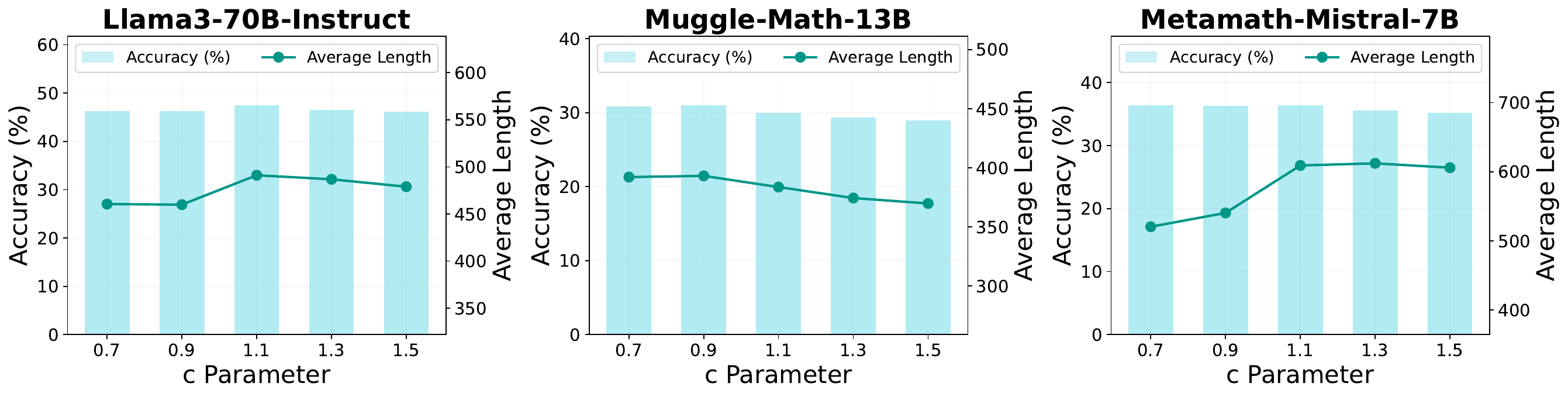}
  \caption{The performance of CoLD PRM under varying values of $c$ across different policy models.}
  \label{fig: hyper-c}

\end{figure}

\begin{figure}[ht]
  \centering

  \includegraphics[width=\textwidth]{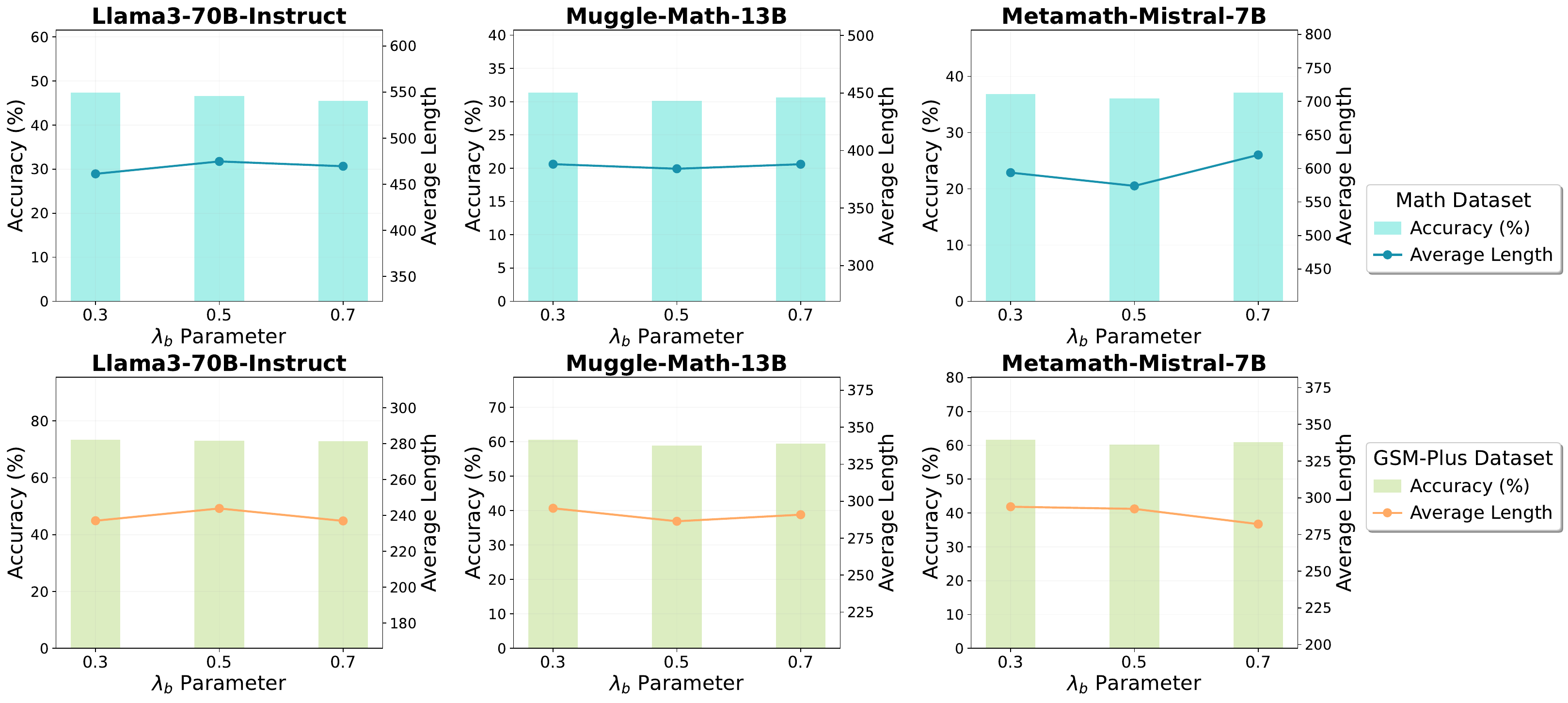}
  \caption{The performance of CoLD PRM under varying values of \(\lambda_b\) across different policy models.}
  \label{fig:hyper-lambda-combined}

\end{figure}

\begin{table}[t]
\centering    

\caption{Effect of different $\alpha$ values on CoLD performance across policy models.}
\label{tab:alpha_ablation}
\resizebox{\textwidth}{!}{
\renewcommand\arraystretch{1}
\begin{tabular}{clcccccc}
\hline
\multirow{2}{*}{\textbf{Policy Model}}&\multirow{2}{*}{\quad\textbf{$\alpha$}} & \multicolumn{2}{c}{\textbf{MATH500}} & \multicolumn{2}{c}{\textbf{GSM-Plus}} & \multicolumn{2}{c}{\textbf{Avg}}  \\ 
\cmidrule(r){3-4} \cmidrule(r){5-6} \cmidrule(r){7-8}
 \multicolumn{2}{c}{} & \textbf{ArithACC(\%)} & \textbf{Length} & \textbf{ArithACC(\%)} & \textbf{Length}& \textbf{ArithACC(\%)} & \textbf{Length}\\ 
\hline 

\multicolumn{1}{c}{\multirow{5}{*}{\makecell{Llama-3-\\70B-Instruct}}}
& 0.1     & 38.8 & 222.6 & 68.6 & 177.4 & 53.7 & 200.0 \\
& 0.01    & 42.4 & 223.5 & 71.1 & 179.6 & 56.8 & 201.6 \\
& 0.001   & 48.6 & 253.5 & 72.6 & 189.7 & 60.6 & 221.6 \\
& 0.0001  & 49.2 & 313.2 & 73.8 & 202.5 & 61.5 & 257.9 \\
& 0.00001 & 47.9 & 350.6 & 73.3 & 210.4 & 60.6 & 280.5 \\
\hline

\multicolumn{1}{c}{\multirow{5}{*}{\makecell{MetaMath-\\Mistral-7B}}}
& 0.1     & 25.0 & 215.3 & 45.5 & 202.9 & 35.3 & 209.1 \\
& 0.01    & 31.2 & 218.9 & 50.2 & 205.3 & 40.7 & 212.1 \\
& 0.001   & 38.8 & 276.29 & 56.9 & 220.2 & 47.9 & 248.2 \\
& 0.0001  & 37.2 & 376.3 & 61.4 & 238.6 & 49.3 & 307.5 \\
& 0.00001 & 36.2 & 393.8 & 59.9 & 259.4 & 48.1 & 326.6 \\
\hline

\multicolumn{1}{c}{\multirow{5}{*}{\makecell{Muggle-\\Math-13B}}}
& 0.1     & 21.4 & 191.1 & 43.7 & 193.1 & 32.6 & 192.1 \\
& 0.01    & 27.0 & 195.9 & 50.0 & 195.8 & 38.5 & 195.9 \\
& 0.001   & 30.2 & 248.4 & 58.5 & 215.8 & 44.4 & 232.1 \\
& 0.0001  & 31.4 & 309.2 & 60.3 & 243.3 & 45.9 & 276.3 \\
& 0.00001 & 30.2 & 357.7 & 59.62 & 263.2 & 44.9 & 310.4 \\
\hline

\end{tabular}
}
\end{table}

\section{Hyperparameter Study}
\label{app:hyper}
\subsection{Effect of $c$ }
We investigate the effect of the hyperparameter \( c \), which controls the strength of the length bias correction in CoLD PRM. We evaluate the performance of CoLD PRM across different values of \( c \in \{0.7, 0.9, 1.1, 1.3, 1.5\} \).

Figure~\ref{fig: hyper-c} illustrates how varying \(c\) affects model accuracy and solution length. The optimal \(c\) varies when different models serve as policy models. For instance, when LLaMA-3-70B-Instruct acts as the policy model, it achieves the highest accuracy at \(c=1.1\); in contrast, when MetaMath-Mistral-7B functions as the policy model, it performs best at \(c=0.7\). Despite these discrepancies, a moderate value like \(c=1.1\) generally delivers strong results across all these policy models. This indicates that CoLD PRM is robust to the selection of \(c\), and using a single \(c\) value can effectively balance accuracy and solution length without the need for extensive per-model tuning.

\subsection{Effect of  $\lambda_b$}
We investigate the effect of the hyperparameter \( \lambda_b\), which controls the strength of the correlation-based regularization in CoLD PRM. We evaluate the performance of CoLD PRM across different values of \( \lambda_b \in \{0.3,0.5,0.7\} \).

Figure~\ref{fig:hyper-lambda-combined} illustrates how varying the value of \( \lambda_b \) affects the model's accuracy and the length of selected solutions. It is evident that different values of \(\lambda_b\) exhibit slight variations in performance across diverse datasets. Specifically, in certain cases, even when the length of the selected solution is relatively short, it may still result in a decline in accuracy. Consequently, an unwavering pursuit of brevity does not necessarily lead to the improvement of performance. Nevertheless, on the whole, CoLD PRM still demonstrates a certain degree of robustness.

\subsection{Effect of $\alpha$}
To further analyze potential over-penalization, we conduct additional experiments on the parameter $\alpha \in \{0.1,0.01,0.001,0.0001,0.00001\}$ across three Policy Model. From the Table ~\ref{tab:alpha_ablation}, we can see that within a reasonable range of $\alpha$, CoLD demonstrates robustness, and the strength of the penalty controlled by $\alpha$ has very limited impact on performance. However, it is also clear that when $\alpha$ applies an excessively strong penalty, over-penalization occurs. This indicates that over-penalization can appear in certain extreme cases.





\begin{table*}[t]
\centering    

\caption{The performance of different models on ProcessBench. 
The best result is given in \textbf{bold}, and the second-best value is \underline{underlined}.}

\label{tab: processbench}
\resizebox{1.0\textwidth}{!}{
\renewcommand\arraystretch{1.1}
\begin{tabular}{clccccccccc}
\hline
\multicolumn{2}{c}{\multirow{2}{*}{Model}} & \multicolumn{2}{c}{GSM8k} & \multicolumn{2}{c}{MATH} & \multicolumn{2}{c}{OlympiadBench} & \multicolumn{2}{c}{OmniMATH} & \multirow{2}{*}{Avg.F1}  \\ 
\cmidrule(r){3-4} \cmidrule(r){5-6} \cmidrule(r){7-8} \cmidrule(r){9-10}
\multicolumn{2}{c}{} & ArithACC & F1 & ArithACC & F1 & ArithACC & F1 & ArithACC & F1 &  \multicolumn{1}{c}{} \\ 
\hline 

\multicolumn{1}{c|}{\multirow{7}{*}{\makecell{Open-source \\ PRM}}}
& CoLD-Base-PRM-7B(Ours) & \underline{72.0} & \underline{68.6} & \textbf{67.3}& \textbf{67.7} & \textbf{54.6} & \textbf{56.0} & \textbf{47.8} & \textbf{51.3} & \textbf{60.9} \\
\multicolumn{1}{c|}{\multirow{4}{*}{}} & Qwen2.5-Math-7B-PRM800K & \textbf{73.5} & 68.2 &\underline{65.1} & \underline{62.6} & \underline{53.2} & \underline{50.7} & \underline{43.4} & \underline{44.3} &  \underline{56.5} \\
\multicolumn{1}{c|}{\multirow{4}{*}{}} & Skywork-PRM-7B & 71.6 & \textbf{70.8} & 54.5 & 53.6 & 25.6 & 22.9 & 23.7 & 21.0 & 42.1 \\
\multicolumn{1}{c|}{\multirow{4}{*}{}} & Skywork-PRM-1.5B & 59.9 & 59.0 & 49.1 & 48.0 & 20.5 & 19.3 & 19.7 & 19.2 & 36.4 \\
\multicolumn{1}{c|}{\multirow{4}{*}{}} & Math-Shepherd-PRM-7B & 58.3 & 47.9 & 45.1 & 29.5 & 39.7 & 24.8 & 34.8 & 23.8 & 31.5 \\
\multicolumn{1}{c|}{\multirow{4}{*}{}} & RLHFlow-PRM-Mistral-8B & 62.3 & 50.4 & 42.1 & 33.4 & 22.3 & 13.8 & 19.1 & 15.8 & 28.4 \\
\multicolumn{1}{c|}{\multirow{4}{*}{}} & RLHFlow-PRM-Deepseek-8B & 56.9 & 38.8 & 45.1 & 33.8 & 26.5 & 16.9 & 23.2 & 16.9 & 26.6 \\

\hline

\multicolumn{1}{c|}{\multirow{17}{*}{\makecell{Language \\ Models \\ as Critic}}}
& QwQ-32B-Preview & \textbf{87.9} & \textbf{88.0} & \textbf{78.5} & \textbf{78.7} & \underline{59.2} & \textbf{57.8} &\textbf{61.1} & \textbf{61.3} & \textbf{71.5} \\
\multicolumn{1}{c|}{\multirow{4}{*}{}} & GPT-4o& 80.2 & 79.2& 63.4 &\underline{63.6} & 50.1&51.4& 50.1 & \underline{53.5} & \underline{61.9} \\
\multicolumn{1}{c|}{\multirow{4}{*}{}} & Qwen2.5-72B-Instruct & 77.9 & 76.2 & \underline{65.4} & 61.8 & \textbf{59.8} & \underline{54.6} & 55.1 & 52.2 &  61.2 \\

\multicolumn{1}{c|}{\multirow{4}{*}{}} & Llama-3.3-70B-Instruct & \underline{83.7} & \underline{82.9} & 63.7 & 59.4 & 54.3 & 46.7 & 51.0 & 43.0 & 58.0 \\
\multicolumn{1}{c|}{\multirow{4}{*}{}} & Qwen2.5-Coder-32B-Instruct & 72.0 & 68.9 & 64.5 & 60.1 & 57.0 & 48.9 & 52.5 & 46.3 & 56.1 \\
\multicolumn{1}{c|}{\multirow{4}{*}{}} & Llama-3.1-70B-Instruct & 75.3 & 74.9 & 52.6 & 48.2 & 50.0 & 46.7 & 43.2 & 41.0 & 52.7 \\
\multicolumn{1}{c|}{\multirow{4}{*}{}} & Qwen2.5-14B-Instruct & 72.3 & 69.3 & 59.2 & 53.3 & 50.2 & 45.0 & 43.5 & 41.3 & 52.2 \\
\multicolumn{1}{c|}{\multirow{4}{*}{}} & Qwen2-72B-Instruct & 67.8 & 67.6 & 52.3 & 49.2 & 43.3 & 42.1 & 39.3 & 40.2 & 49.8 \\
\multicolumn{1}{c|}{\multirow{4}{*}{}} & Qwen2.5-32B-Instruct & 70.6 & 65.6 & 61.9 & 53.1 & 53.5 & 40.0 & 47.7 & 38.3 & 49.3 \\
\multicolumn{1}{c|}{\multirow{4}{*}{}} & Qwen2.5-Math-72B-Instruct & 70.3 & 65.8 & 59.6 & 52.1 & 56.1 & 32.5 & 55.1 & 31.7 & 45.5 \\
\multicolumn{1}{c|}{\multirow{4}{*}{}} & Qwen2.5-Coder-14B-Instruct & 61.9 & 50.1 & 54.2 & 39.9 & 51.4 & 34.0 & \underline{55.6} & 27.3 & 37.8 \\
\multicolumn{1}{c|}{\multirow{4}{*}{}} & Qwen2.5-7B-Instruct & 37.8 & 36.5 & 36.9 & 36.6 & 29.9 & 29.7 & 27.3 & 27.4 & 32.6 \\
\multicolumn{1}{c|}{\multirow{4}{*}{}} & Meta-Llama-3-70B-Instruct & 62.4 & 52.2 & 48.3 & 22.8 & 46.2 & 21.2 & 44.8 & 20.0 & 29.1 \\
\multicolumn{1}{c|}{\multirow{4}{*}{}} & Qwen2.5-Math-7B-Instruct & 54.4 & 26.8 & 50.3 & 25.7 & 43.1 & 14.2 & 41.6 & 12.7 & 19.9 \\
\multicolumn{1}{c|}{\multirow{4}{*}{}} & Qwen2-7B-Instruct & 25.1 & 8.4 & 20.4 & 19.0 & 16.1 & 14.7 & 13.8 & 12.1 & 13.6 \\
\multicolumn{1}{c|}{\multirow{4}{*}{}} & Meta-Llama-3-8B-Instruct & 27.1 & 13.1 & 17.3 & 13.8 & 14.2 & 4.8 & 19.7 & 12.6 & 11.1 \\
\multicolumn{1}{c|}{\multirow{4}{*}{}} & Qwen2.5-Coder-7B-Instruct & 49.1 & 14.3 & 46.3 & 6.5 & 47.2 & 4.1 & 48.9 & 1.8 & 6.7 \\
\multicolumn{1}{c|}{\multirow{4}{*}{}} & Llama-3.1-8B-Instruct & 27.3 & 10.9 & 20.5 & 5.1 & 16.0 & 2.8 & 15.0 & 1.6 & 5.1 \\

\hline
\end{tabular}

}
\end{table*}
\section{Other Benchmark}
\label{app: Other Benchmark}
To better evaluate the capabilities of our model, we assess it on ProcessBench~\citep{zheng2024processbench}—a public benchmark for Process Reward Models (PRMs). The goal of this evaluation is to determine whether the PRM can identify the first erroneous step in the reasoning process. ProcessBench partitions the dataset into two subsets: one containing samples with incorrect final answers and the other with correct final answers. It then computes the harmonic mean of the accuracies achieved on these two subsets to obtain the final F1-score. 

As shown in the Table~\ref{tab: processbench}, our CoLD PRM outperforms nearly all open-source PRM baselines across all datasets. The performance advantage is particularly notable on the more challenging datasets, including OlympiadBench and OmniMATH, demonstrating that our CoLD PRM maintains strong generalization over OOD datasets. 

When comparing models across different scales, CoLD still maintains strong competitiveness. In terms of overall performance, it is even able to outperform some larger-scale counterpart models.

\begin{table*}[ht]
\centering

\caption{Evaluation Results of CoLD Using Advanced Large-Scale Policy Models on MATH500.}

\renewcommand\arraystretch{1}
\begin{tabular}{clcc}
\hline
Policy Model                                                                                                 & PRM model      & ArithACC(\%) & Length \\ \hline
\multicolumn{1}{c|}{\multirow{2}{*}{\begin{tabular}[c]{@{}c@{}}Qwen3-Next-80B\\ -A3B-Instruct\end{tabular}}} & Vanilla Base PRM    & 85.2         & 1050.6 \\
\multicolumn{1}{c|}{}                                                                                        & CoLD Base PRM(Ours) & 90.4         & 754.5  \\ \hline
\multicolumn{1}{c|}{\multirow{2}{*}{Deepseek-v3.1}}                                                          & Vanilla Base PRM    & 89.6         & 953.9  \\
\multicolumn{1}{c|}{}                                                                                        & CoLD Base PRM(Ours) & 91.6         & 646.2  \\ \hline
\end{tabular}
\label{tab: advance}
\end{table*}

\section{Evaluation Using Stronger Policy Model}
\label{app:stronger model}
In addition to the policy models reported in the main paper, we further include Qwen3-Next-80B-A3B-Instruct~\cite{yang2025qwen3} and DeepSeek-v3.1~\cite{liu2024deepseek} in our evaluation. These models represent state-of-the-art large-scale language models that have demonstrated substantially stronger reasoning capabilities compared with the earlier baselines.

As shown in the Table~\ref{tab: advance}, our CoLD framework continues to deliver clear and consistent improvements even when applied to these more advanced LLMs. This confirms that CoLD is not merely compensating for weaknesses in smaller models, but instead provides generalizable gains that hold across diverse policy models, including the latest high-performing LLMs.

\begin{figure}[ht]
  \centering

  \includegraphics[width=\textwidth]{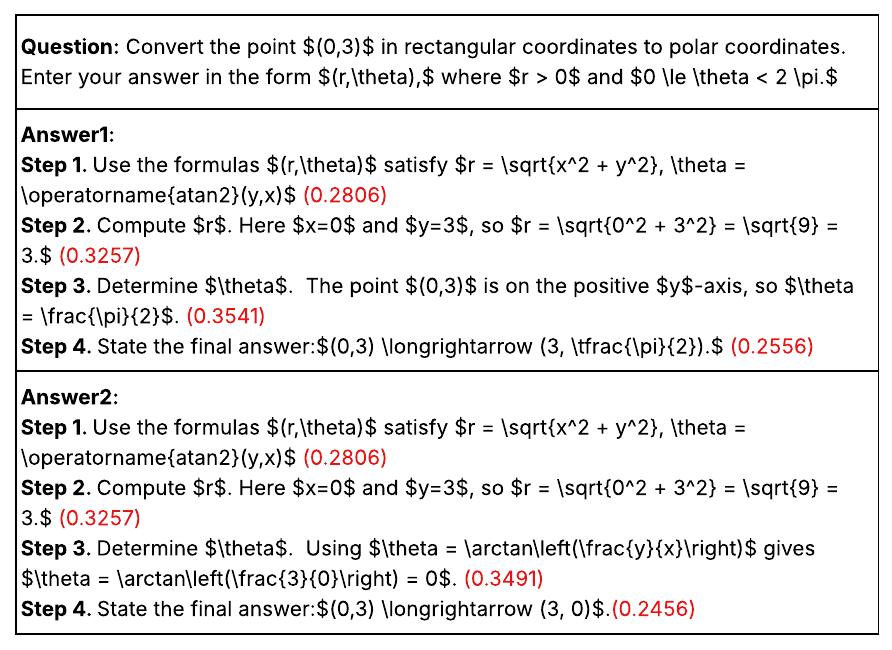}
  \caption{Case study of the Bias Estimator outputs, where the red numbers in parentheses at the end of each step denote the predicted scores.}
  \label{fig:case study}

\end{figure}

\begin{figure}[ht]
  \centering

  \includegraphics[width=\textwidth]{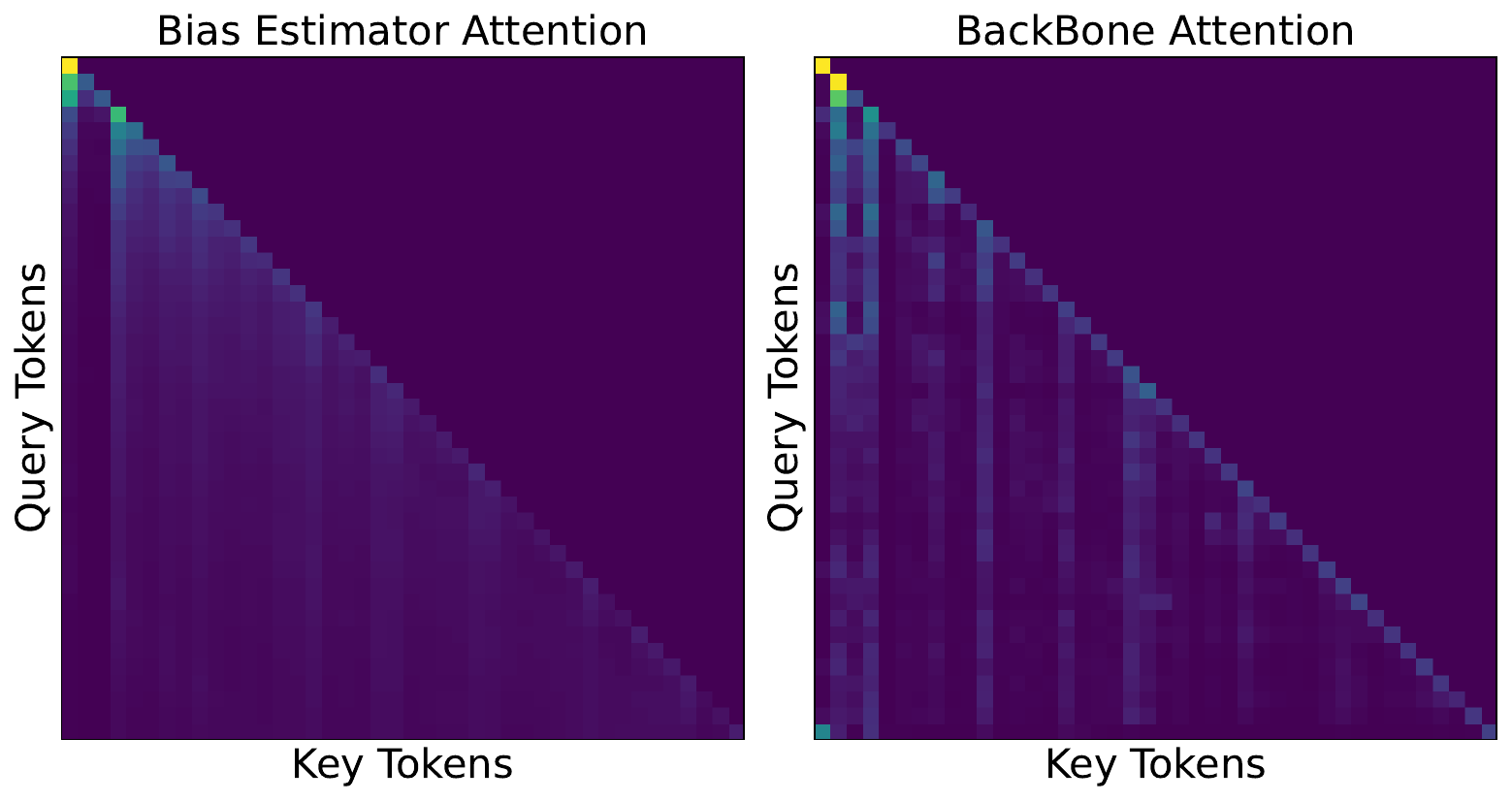}
  \caption{Attention visualizations for Sentence A in the backbone model and the Bias Estimator.}
  \label{fig:backbone compare}

\end{figure}

\begin{figure}[ht]
  \centering

  \includegraphics[width=\textwidth]{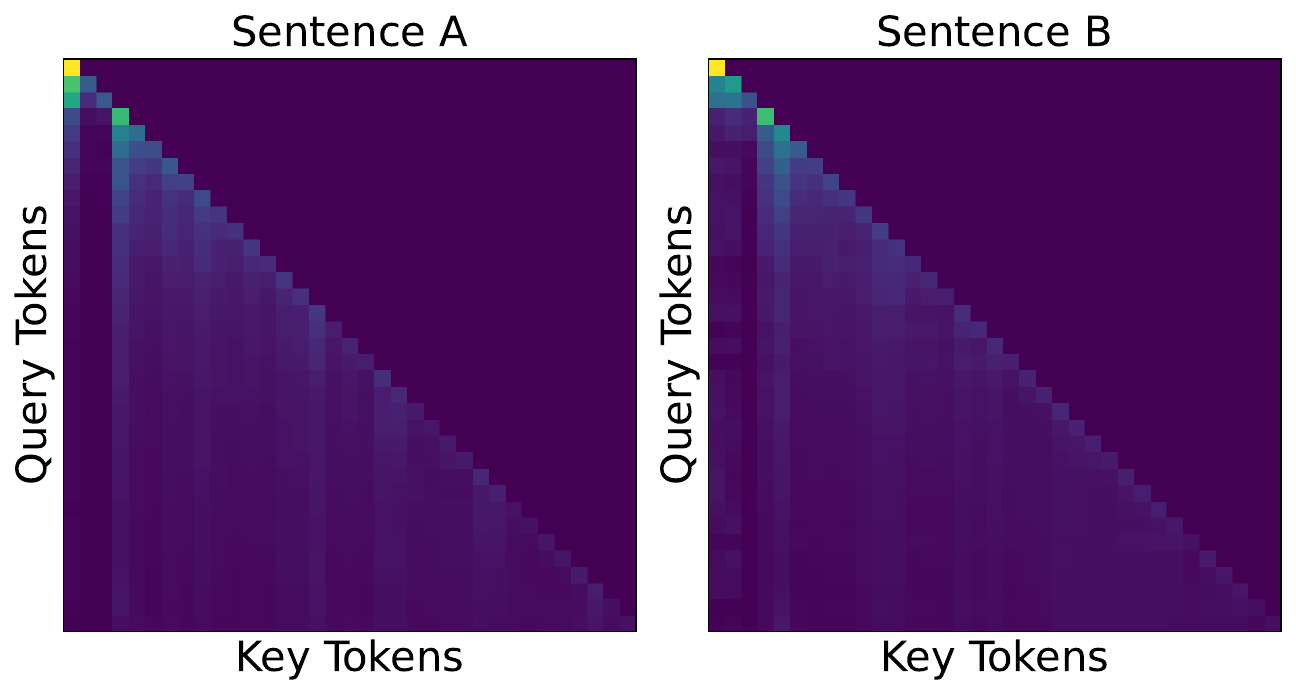}
  \caption{Attention visualizations for Sentence A and Sentence B, which share identical length but differ in semantics, under the Bias Estimator.}
  \label{fig:same attention}

\end{figure}

\section{Bias Estimator Interpretability}
\label{app:Interpretability}
To better understand what the Bias Estimator learns and how it makes predictions, we conduct interpretability analyses focusing on its sensitivity to input length versus semantic content. Through a combination of output case studies and attention-based comparisons, we aim to clarify whether the Bias Estimator captures spurious surface-level cues, specifically length, rather than the underlying meaning of the input.
\subsection{Case Study}
To illustrate the interpretability of the Bias Estimator, we present a case study demonstrating its output behavior. As shown in Figure~\ref{fig:case study}, the Bias Estimator responds more strongly to length than to semantic content. For example, in Step 3 of the two answers displayed, although their semantic meaning differs and one step is correct while the other is incorrect, the Bias Estimator assigns similar scores because it is primarily driven by step length rather than semantics. 

\subsection{Attention Visualization}
\begin{itemize}
    \item Sentence A: Compute $r$. Here $x=0$ and $y=3$, so $r = \sqrt{0^2 + 3^2} = \sqrt{9} = 3$.
    \item Sentence B: Analyze $k$. If a machine runs 12 cycles in 4 seconds, then $k=12/4=3$ .
\end{itemize}
To examine whether the Bias Estimator relies on length rather than semantic information, we conduct two comparative attention analyses. First, we feed Sentence A into both the Bias Estimator and the backbone model and compare their attention distributions. As shown in Figure ~\ref{fig:backbone compare}, while the backbone exhibits semantically focused attention peaks, the Bias Estimator assigns attention almost uniformly across tokens, without concentrating on semantic-bearing words. Second, we input Sentence A and Sentence B, which share the same length but differ completely in semantics, into the Bias Estimator. As illustrated in Figure ~\ref{fig:same attention}, their attention maps remain nearly indistinguishable, indicating that the attention pattern remains stable when semantics change while length is kept constant. Together, these results reinforce that the Bias Estimator’s attention primarily encodes length-related cues rather than semantic information.

\section{Limitation}
\label{app:Limitation}
We identify two primary limitations in our approach. First, the framework relies on heuristic noise injection to approximate counterfactual scenarios. A true counterfactual requires a reasoning step strictly stripped of all semantic content while retaining its exact surface length. Since this ideal state is fundamentally unobservable, the Bias Estimator depends on perturbation approximations. Consequently, the estimated causal effect of length remains a proxy, and the bias estimation could potentially be influenced by residual semantic leakage depending on the noise configuration. Second, the framework introduces additional hyperparameters, specifically the bias correction strength ($c$), the correlation regularization weight ($\lambda_b$), and the length penalty ($\alpha$). The presence of these variables intrinsically increases the search space for hyperparameter tuning. This expanded search space could complicate deployment and scaling across highly diverse or unseen policy models in practical applications.

\section{Broader Impacts}
\label{app:Broader impacts}
Our research addresses the phenomenon of length bias in Process Reward Models (PRMs), yielding several positive societal impacts. By explicitly disentangling true semantic correctness from superficial verbosity, our CoLD framework promotes fairer and more reliable automated evaluation for Large Language Models (LLMs). This mitigates the risk of reward hacking, where models are incentivized to generate redundant or hallucinatory text simply to attain higher scores. Additionally, by encouraging more concise and precise reasoning trajectories, our approach can help reduce the computational overhead and carbon footprint associated with LLM inference, which aligns with the objectives of environmentally sustainable AI.

Regarding potential negative impacts, our work shares the dual-use risks inherent to any foundational advancement in AI reasoning. While our framework is designed to improve logical correctness in mathematical problem-solving, enhanced reward modeling techniques could theoretically be repurposed by bad actors to optimize models for generating sophisticated deceptive logic or malicious code. Since our focus is on foundational methodology, mitigating these downstream risks relies on the broader AI community to implement robust safety guardrails and usage monitoring at the final deployment stage.

\section*{Large Language Models (LLMs) Usage}
In this work, large language models (LLMs) were used solely for text polishing and language refinement. They were not involved in the design of the methodology, implementation, analysis, or the generation of experimental results. All technical contributions and research findings are entirely the work of the authors.

\newpage
\section*{NeurIPS Paper Checklist}

\begin{enumerate}

\item {\bf Claims}
    \item[] Question: Do the main claims made in the abstract and introduction accurately reflect the paper's contributions and scope?
    \item[] Answer: \answerYes{} 
    \item[] Justification: The main claims made in the abstract and introduction accurately reflect the paper’s contributions and scope.

\item {\bf Limitations}
    \item[] Question: Does the paper discuss the limitations of the work performed by the authors?
    \item[] Answer: \answerYes{} 
    \item[] Justification: The paper discusses the limitations of our work. Check Appendix \ref{app:Limitation}.

\item {\bf Theory assumptions and proofs}
    \item[] Question: For each theoretical result, does the paper provide the full set of assumptions and a complete (and correct) proof?
    \item[] Answer: \answerYes{} 
    \item[] Justification: For each theoretical result, the paper provides the full set of assumptions and a complete (and correct) proof.

    \item {\bf Experimental result reproducibility}
    \item[] Question: Does the paper fully disclose all the information needed to reproduce the main experimental results of the paper to the extent that it affects the main claims and/or conclusions of the paper (regardless of whether the code and data are provided or not)?
    \item[] Answer: \answerYes{} 
    \item[] Justification:  Check experimental setups Section~\ref{exp:setup}, Appendix~\ref{app:Experiment Details}.

\item {\bf Open access to data and code}
    \item[] Question: Does the paper provide open access to the data and code, with sufficient instructions to faithfully reproduce the main experimental results, as described in supplemental material?
    \item[] Answer: \answerYes{} 
    \item[] Justification: All code and data necessary to reproduce our results is included in supplemental material.

\item {\bf Experimental setting/details}
    \item[] Question: Does the paper specify all the training and test details (e.g., data splits, hyperparameters, how they were chosen, type of optimizer) necessary to understand the results?
    \item[] Answer: \answerYes{} 
    \item[] Justification:  Check experimental setups Section~\ref{exp:setup}, Appendix~\ref{app:Experiment Details}.

\item {\bf Experiment statistical significance}
    \item[] Question: Does the paper report error bars suitably and correctly defined or other appropriate information about the statistical significance of the experiments?
    \item[] Answer: \answerYes{} 
    \item[] Justification: We have included statistical significance markers directly in the main performance table to denote significant improvements over the baselines.

\item {\bf Experiments compute resources}
    \item[] Question: For each experiment, does the paper provide sufficient information on the computer resources (type of compute workers, memory, time of execution) needed to reproduce the experiments?
    \item[] Answer: \answerYes{} 
    \item[] Justification:  Check experimental setups Section~\ref{exp:setup}, Appendix~\ref{app:Experiment Details}.
    
\item {\bf Code of ethics}
    \item[] Question: Does the research conducted in the paper conform, in every respect, with the NeurIPS Code of Ethics \url{https://neurips.cc/public/EthicsGuidelines}?
    \item[] Answer: \answerYes{} 
    \item[] Justification:  The research conducted in the paper does confirm, in every respsect, with the NeurIPS Code of Ethics.

\item {\bf Broader impacts}
    \item[] Question: Does the paper discuss both potential positive societal impacts and negative societal impacts of the work performed?
    \item[] Answer: \answerYes{} 
    \item[] Justification: We discuss the broader impacts in the appendix~\ref{app:Broader impacts}. 
    
\item {\bf Safeguards}
    \item[] Question: Does the paper describe safeguards that have been put in place for responsible release of data or models that have a high risk for misuse (e.g., pre-trained language models, image generators, or scraped datasets)?
    \item[] Answer: \answerNA{} 
    \item[] Justification: The paper poses no such risks.

\item {\bf Licenses for existing assets}
    \item[] Question: Are the creators or original owners of assets (e.g., code, data, models), used in the paper, properly credited and are the license and terms of use explicitly mentioned and properly respected?
    \item[] Answer: \answerYes{} 
    \item[] Justification: All datasets and code are open-source and follow the license of the original work. Check experimental setups Section~\ref{exp:setup}, Appendix~\ref{app:Experiment Details}.

\item {\bf New assets}
    \item[] Question: Are new assets introduced in the paper well documented and is the documentation provided alongside the assets?
    \item[] Answer: \answerNA{} 
    \item[] Justification: The paper does not release new assets.

\item {\bf Crowdsourcing and research with human subjects}
    \item[] Question: For crowdsourcing experiments and research with human subjects, does the paper include the full text of instructions given to participants and screenshots, if applicable, as well as details about compensation (if any)? 
    \item[] Answer: \answerNA{} 
    \item[] Justification: We did not conduct crowdsourcing experiments; human evaluation was strictly limited to verifying the consistency between the rewritten and original solution steps.

\item {\bf Institutional review board (IRB) approvals or equivalent for research with human subjects}
    \item[] Question: Does the paper describe potential risks incurred by study participants, whether such risks were disclosed to the subjects, and whether Institutional Review Board (IRB) approvals (or an equivalent approval/review based on the requirements of your country or institution) were obtained?
    \item[] Answer: \answerNA{} 
    \item[] Justification: The paper does not involve crowdsourcing nor research with human subjects.

\item {\bf Declaration of LLM usage}
    \item[] Question: Does the paper describe the usage of LLMs if it is an important, original, or non-standard component of the core methods in this research? Note that if the LLM is used only for writing, editing, or formatting purposes and does \emph{not} impact the core methodology, scientific rigor, or originality of the research, declaration is not required.
    \item[] Answer: \answerNA{} 
    \item[] Justification: The core method development in this research does not involve LLMs as any
important, original, or non-standard components.

\end{enumerate}

\end{document}